\documentclass[runningheads]{llncs}

\usepackage{accv}

\usepackage{accvabbrv}
\usepackage{multirow}
\usepackage{multicol}
\usepackage{graphicx}
\usepackage{booktabs}
\usepackage{subcaption}
\usepackage{diagbox} 
\usepackage{wrapfig}

\usepackage[accsupp]{axessibility}  

\usepackage{hyperref}

\usepackage{orcidlink}

\begin{document}

\title{Beyond Coarse-Grained Matching in \\Video-Text Retrieval} 


\author{Aozhu Chen\inst{1} \thanks{Work was part of an internship at University of Amsterdam}
\and
Hazel Doughty\inst{2}
\and
Xirong Li\inst{1}
\and
Cees G. M. Snoek\inst{3}
}


\institute{Renmin University of China\and Leiden University \and University of Amsterdam}

\maketitle

\vspace{-1.8em}
\begin{abstract}
%
Video-text retrieval has seen significant advancements, yet the ability of models to discern subtle differences in captions still requires verification. In this paper, we introduce a new approach for fine-grained evaluation. Our approach can be applied to existing datasets by automatically generating hard negative test captions with subtle single-word variations across nouns, verbs, adjectives, adverbs, and prepositions. We perform comprehensive experiments using four state-of-the-art models across two standard benchmarks (MSR-VTT and VATEX) and two specially curated datasets enriched with detailed descriptions (VLN-UVO and VLN-OOPS), resulting in a number of novel insights: 1) our analyses show that the current evaluation benchmarks fall short in detecting a model's ability to perceive subtle single-word differences, 2) our fine-grained evaluation highlights the difficulty models face in distinguishing such subtle variations. To enhance fine-grained understanding, we propose a new baseline that can be easily combined with current methods. Experiments on our fine-grained evaluations demonstrate that this approach enhances a model's ability to understand fine-grained differences.


 \vspace{-0.5em} 
\end{abstract}

\vspace{-2.1em}
\section{Introduction}
\vspace{-1em}
\label{sec:intro}
\begin{wrapfigure}{r}{0.36\textwidth}
    \centering
   \vspace{-4.5em}
    \begin{subfigure}[b]{\linewidth}
        \includegraphics[width=\textwidth]{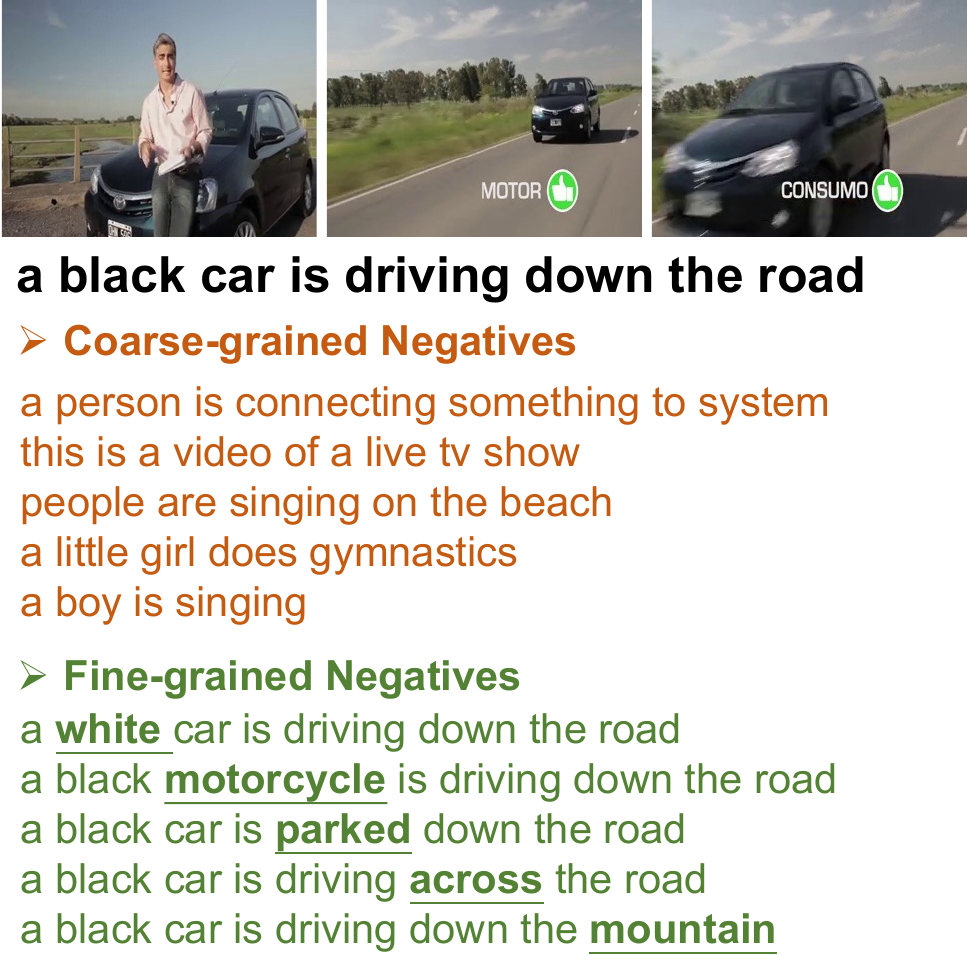}
        \label{fig:sub2}
    \end{subfigure}
    \vspace{-3em}
    \caption{\textbf{Coarse vs. Fine-Grained Video Retrieval.} }
    \vspace{-2em}
     \label{fig:captions-example}
\end{wrapfigure}
In recent years, video-text retrieval has seen huge improvements~\cite{bmvc2018vsepp,LiXirong2019W2VVPP,wray2019JPoSE,chen2020hgr,hu2021lightweight_LAFF,tpami22-duale,bain2021frozen,xclip,ts2net,wang2023ucofia,Tian2024CVPR}. This increased success is bolstered by large-scale web-scraped vision-language datasets~\cite{doughty2022how} and the ability to bootstrap training with more
readily available image-text pairs~\cite{sharma2018gcc}. Popular video-text retrieval benchmarks~\cite{chen2011msvd,xu2016msr,wang2019vatex,tgif2015,activitynetcap,didemo} have paved the way for this success, with models often able to recall the correct caption from thousands in the top 10, 5 or even first result. Although impressive, this is only a step towards fully understanding the correspondence between video and text. The collection process of current benchmarks results in a diverse set of videos and corresponding captions which are distinct from each other. 
This means that current benchmarks only evaluate coarse-grained video retrieval and that we do not know the fine-grained capabilities of current models. 
As shown in \cref{fig:captions-example}, the ground-truth caption (black) for the given video is distinctly different from the other captions in the test set (coarse-grained negatives in brown). This makes it relatively easy for the model to distinguish between these captions. We refer to matching and retrieving such videos and texts as coarse-grained evaluation. However, it is currently unknown whether a model can discern the subtle single-word differences between the fine-grained negative captions (green) and the original caption, which is challenging and critical for accurate video-text alignment.

In this paper, we propose to go beyond coarse-grained matching in video-text retrieval and focus on fine-grained differences in captions. We create a new fine-grained evaluation approach that can be used with existing datasets by automatically generating new hard negative test captions with subtle single-word variations across different parts-of-speech. We find that current models struggle to distinguish such subtle variations, particularly in prepositions and adverbs. The challenge in addressing this is not only to make a model understand variations in different parts-of-speech but to gain this understanding while maintaining a model's impressive coarse-grained abilities. To take a first step towards rectifying this, we propose a plug-in method that balances coarse- and fine-grained objectives with phrase-level negatives and fine-grained prompting.

Our contributions can be summarized as: (i) we highlight the coarse-grained nature of current video-retrieval evaluation, (ii) we propose a new fine-grained evaluation approach to analyze models' sensitivity to individual word variations in different parts-of-speech, (iii) we investigate current models using our proposed evaluation and find they lack fine-grained understanding, and (iv) we propose a simple method which can be easily combined with existing approaches to learn representations suitable for both coarse- and fine-grained retrieval.

\vspace{-1em}
\section{Related Work}
\vspace{-0.5em}
\label{sec:related-work}
\noindent\textbf{Coarse-Grained Video-Text Retrieval.}  
Video-text retrieval is an important topic in the vision-language domain that has garnered much attention~\cite{yu2018joint, croitoru2021teachtext, yang2021taco, wang2021t2vlad, chen2020uniter, alayrac2022flamingo, zellers2022merlot, chen2020fine, wang2023video, lu2023uniadapter, cheng2023cico, shu2024mac}. The aim is to retrieve the correct video given a text query. Thus, it is crucial to align the features of videos to their related texts. Early works in video-text retrieval~\cite{torabi2016learning, yu2018joint, yu2016video} design different fusion mechanisms for pre-extracted video and text features. Later, Clip-BERT~\cite{lei2021less} enabled end-to-end training with a sparse sampling strategy on video data in addition to image-text pretraining. Frozen~\cite{bain2021frozen} also uses image-text pretraining with a curriculum learning schedule. Several works~\cite{luo2022clip4clip, bain2022clip, fang2021clip2video, buch2022revisiting, jiang2022cross} instead build on CLIP~\cite{radford2021learning}, 
repurposing it for video-text retrieval. However, these works can only understand coarse-grained differences between text and video; they fail to recognize the subtle one-word caption differences investigated in this paper.
\newline
\noindent\textbf{Token-Level Similarity in Video-Text Retrieval.} 
Several works aim to make the understanding of visual and text correspondence more fine-grained. One category of methods enhances retrieval performance by designing different granularities of cross-modal similarity~\cite{xclip, wang2023ucofia, zou2022tokenflow, lee2018stacked, messina2021fine, yao2021filip, gou2023leveraging}. 
For example, TS2Net~\cite{ts2net} designs a cross-
modal alignment module between caption and frame features, rather than only the video-level features. X-CLIP \cite{xclip} calculates similarities between video-sentence, video-word, sentence-frame, and frame-word, using varying levels of similarity to allow the model to perceive multi-granularity alignments. Similarly, UCoFiA \cite{wang2023ucofia} uses coarse-to-fine similarity alignment, going a step further with patch-word similarity. While achieving impressive results on current video-to-text benchmarks, we show in this paper that these works still struggle to understand fine-grained single-word differences in video captions.
\newline
\noindent\textbf{Parts-of-Speech in Video-Text Retrieval.} Other methods aim for fine-grained understanding by decomposing the input caption into different parts-of-speech~\cite{chen2020hgr, wray2019JPoSE, doughty2020action, bagad2023test, momeni2023verbs, wang2022negation}. For instance, HGR \cite{chen2020hgr} uses hierarchical graph reasoning on verbs, noun phrases, and complete sentences to model a semantic graph of events, actions, entities, and their relationships.
JPoSE \cite{wray2019JPoSE} learns separate embedding spaces for verbs and nouns before combining these into a fine-grained action embedding. 
ActionInVerb \cite{momeni2023verbs} improves verb understanding by leveraging pretrained large language models to generate hard negatives which only change the verbs in sentences.  Several works also go beyond verb-noun understanding in video-text retrieval and examine individual parts-of-speech with adverbs~\cite{doughty2022how, doughty2020action}, temporal prepositions~\cite{bagad2023test} and negation~\cite{wang2022negation}. 
Our work also focuses on fine-grained video-text retrieval with different parts-of-speech, however unlike prior work, it goes beyond investigating verb-noun pairs or single parts-of-speech.
\newline
\noindent\textbf{Analysis of Vision-Language Models.}
Several studies have investigated the comprehension of fine-grained information in vision-language models~\cite{shekhar2017foil, hendricks2021svoprob, madasu2024, c-iclr23-whenandwhy, e-eccv24-longclip, d-arxiv22-vlchecklist, wray2021semantic, chan2022s, fan2021negative}. Foilit \cite{shekhar2017foil} creates an erroneous description (`foil'), by replacing nouns in an image caption with similar but incorrect words, testing models with foil classification, foil word detection, and foil word correction. 
SVO-Probes \cite{hendricks2021svoprob} assess image-text models' understanding of verbs by evaluating images related or unrelated to specific sentences with subject, verb, and object triplets. 
Madasu \etal \cite{madasu2024} investigate compositional and semantic understanding in video retrieval by removing, shifting, or replacing, verbs and nouns in the ground-truth captions.  Similarly, other works \cite{c-iclr23-whenandwhy, e-eccv24-longclip} also highlight the lack of compositional understanding in image-text models. 
VL-Checklist \cite{d-arxiv22-vlchecklist} instead tests VLMs' understanding of objects, attributes, and relations, while Wray \etal~\cite{wray2021semantic} focus on models' understanding of different captions with similar semantic meanings. Different to these prior works, this paper explores video-text retrieval models' understanding of nouns, adjectives, verbs, adverbs, and prepositions, focusing specifically on whether models can accurately understand and respond to subtle one-word changes in language.



\vspace{-1em}
\section{Shortcomings of Current Video Retrieval Benchmarks}
\vspace{-0.5em}
\label{sec:coarse-metric}
In this section, we highlight the coarse-grained nature of the current video-text retrieval paradigm by examining current datasets and the captions they contain.

\noindent\textbf{Definition \& Notation. }
Video-text retrieval models aim to learn representations to determine the relevance of texts $T$ to videos $V$. Specifically, given a video $v\in V$, the aim is to rank the set of texts $T$ according to their relevance to $v$ and vice versa. 
What is learned by a representation thus depends heavily on the content of the captions labeled as relevant and irrelevant to each video. Similarly, the insights gained from model evaluation hinges on the captions present in the test set. If the negative captions correspond to clear visual differences, we do not know if a model can distinguish fine-grained differences. Similarly, if captions can be distinguished by one part-of-speech alone, \eg nouns, we do not know if a model can understand the visual appearance of different parts-of-speech. 
\begin{table}[t]
\caption{\textbf{Details of Benchmarks in Text-to-Video Retrieval}. We show the number of videos and captions in the train and test sets, alongside the proportions of captions containing nouns, verbs, adjectives, adverbs, and prepositions.}
\vspace{-1em}
\label{tab:benchmarks}
\resizebox{\linewidth}{!}{
\begin{tabular}{lrrrrrrrrrrrrrrrc}
\midrule
  & & &\multicolumn{2}{c}{Training Set} & & \multicolumn{2}{c}{Test Set} & & & \multicolumn{5}{c}{Proportions of Captions} \\
  \cmidrule{4-5} \cmidrule{7-8} \cmidrule{11-15}
\multicolumn{1}{l}{\multirow{-2}{*}{Dataset}} & \multirow{-2}{*}{Source}&  & \#Video & \multicolumn{1}{c}{\multirow{1}{*}{\#Caption}} & & \multicolumn{1}{c}{\#Video} & \multicolumn{1}{c}{\#Caption} & & \multirow{-2}{*}{\#Vocab} & \multicolumn{1}{c}{Noun} & \multicolumn{1}{c}{Verb} & \multicolumn{1}{c}{Adj} & \multicolumn{1}{c}{Adv} & \multicolumn{1}{c}{Prep}  & & \multirow{-2}{*}{Eval. Metrics}\\ \midrule 
YouCook2 \cite{Zhou2018youcook2} & YouTube &&13.8K& 13.8K &  & 3.3K & 3.3K & & 2.3K  & 1.00 & 0.95 & 0.33 & 0.10 & 0.75 && Recall@k, Av. Rank\\
DiDeMo \cite{didemo} & Flicker& & 10.6K& 41.2K &  & 1.0K & 4.0K & & 6.8K & 1.00 & 0.88 & 0.44 & 0.24 & 0.67 && Recall@k \\
ActivityNet \cite{activitynetcap} & YouTube& & 10.9K & 10.9K & & 4.9K & 4.9K & & 13.0K  & 0.99 & 0.99 & 0.42 & 0.33 & 0.83 & & Recall@k \\
VATEX \cite{wang2019vatex} &  YouTube & &28.9K&  289.9K & & 1.5K & 15.0K & & 29.1K &  1.00 & 0.99 & 0.50 & 0.25 & 0.89  & & Recall@k, Av. Rank\\
MSR-VTT \cite{xu2016msr} & YouTube &&9.0K & 180.0K & & 1.0K & 20.0K & & 25.7K  & 1.00 & 0.92 & 0.36 & 0.09 & 0.68   & & Recall@k, Av. Rank\\
MSVD  \cite{chen2011msvd} & YouTube && 1.3K& 52.9K & & 670 & 27.8K & & 12.9K  & 1.00 & 0.96 & 0.18 & 0.05 & 0.46  & & Recall@k, Av. Rank\\
TGIF \cite{tgif2015} & Tumblr && 89.4K& 90.4K &  & 11.3K  & 34.1K & & 11.4K &  1.00 & 0.99 & 0.43 & 0.20 & 0.78 &&Recall@k, Av. Rank \\

\bottomrule
\end{tabular}
}
\vspace{-0.9em}
\end{table}

\noindent\textbf{Datasets. }
In \cref{tab:benchmarks} we show the statistics of commonly used video-text retrieval benchmarks. We order these by the size of the test set. A larger test set is generally more challenging as a model not only needs to distinguish between more items but there is an increased chance of having fine-grained differences between items.
Most datasets originate from YouTube or other websites such as Flickr and Tumblr and are annotated after the fact via crowd-sourcing. During this annotation, an annotator observes a single video and is tasked with creating a relevant caption. This caption is not created by considering the other videos or captions present in the dataset. All datasets are evaluated using Recall@k or Avg. Rank where the relevant caption(s) should be retrieved in the top k results or higher than the irrelevant captions sourced from other videos. The granularity of this evaluation thus relies on the contents of the test set captions which rarely have subtle differences between each other due to the collection process.
The proportions of captions with each word type in \cref{tab:benchmarks} also suggest these datasets do not test fine-grained understanding. Subjects, objects (nouns), and actions (verbs) are the most central elements, while the manner of actions (adverbs) is less common in captions. 

\begin{figure}[tb]
  \centering
  \includegraphics[width = \linewidth]{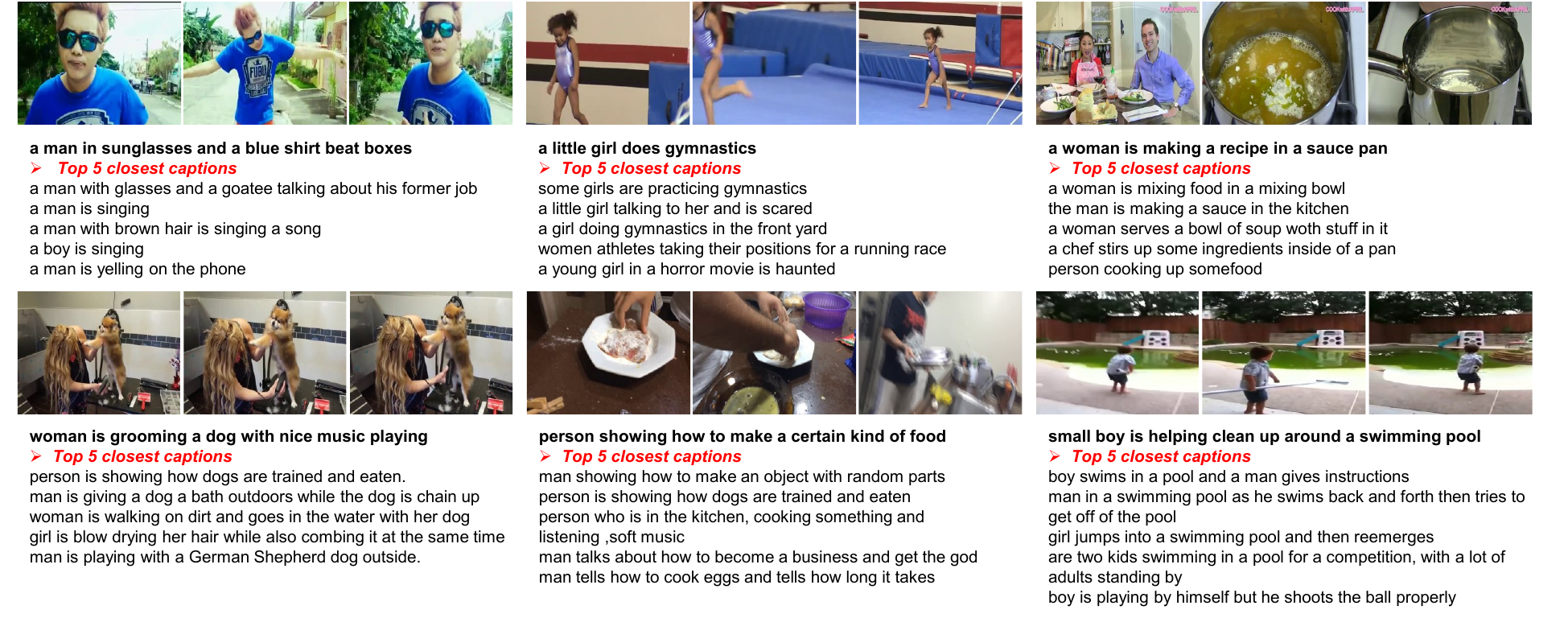}
  \vspace{-2em}
  \caption{\textbf{Qualitative Analysis on MSR-VTT \& VATEX.} We show ground-truth video-text pairs alongside the 5 most similar captions to the ground-truth according to CLIP. We see that captions are coarse-grained, \ie there are multiple difference conceptual differences between captions which mostly relate to the nouns present.}
    \label{fig:qual_anal}
  \vspace{-2em}
\end{figure}

\noindent\textbf{Qualitative Analysis. }
To investigate how fine-grained the captions in these datasets are we take a closer look at example captions in \cref{fig:qual_anal}. Here, we show randomly selected videos from MSR-VTT \cite{xu2016msr} and VATEX \cite{wang2019vatex} alongside the ground-truth caption and the top-5 closest captions in the test set according to CLIP\cite{2021clip_icml} similarity. We see that the closest captions in the test set differ by multiple concepts \eg `a little girl does gymnastics' vs. `women athletes taking their position for a running race'. This suggests that these datasets inadequately test the fine-grained capabilities of models. Furthermore, most differences are due to different nouns and verbs, none of the examples share the same noun and verb, differing only in adjectives, adverbs, and prepositions.

\noindent\textbf{Quantitative Analysis. }
To verify our observations from ~\cref{fig:qual_anal} hold true for the rest of the video-text pairs in these datasets, we conduct statistical analysis  
\begin{wrapfigure}{r}{0.36\textwidth}
    \centering
    \vspace{-0.2em}
    \begin{subfigure}[b]{\linewidth}
        \includegraphics[width=0.95\textwidth]{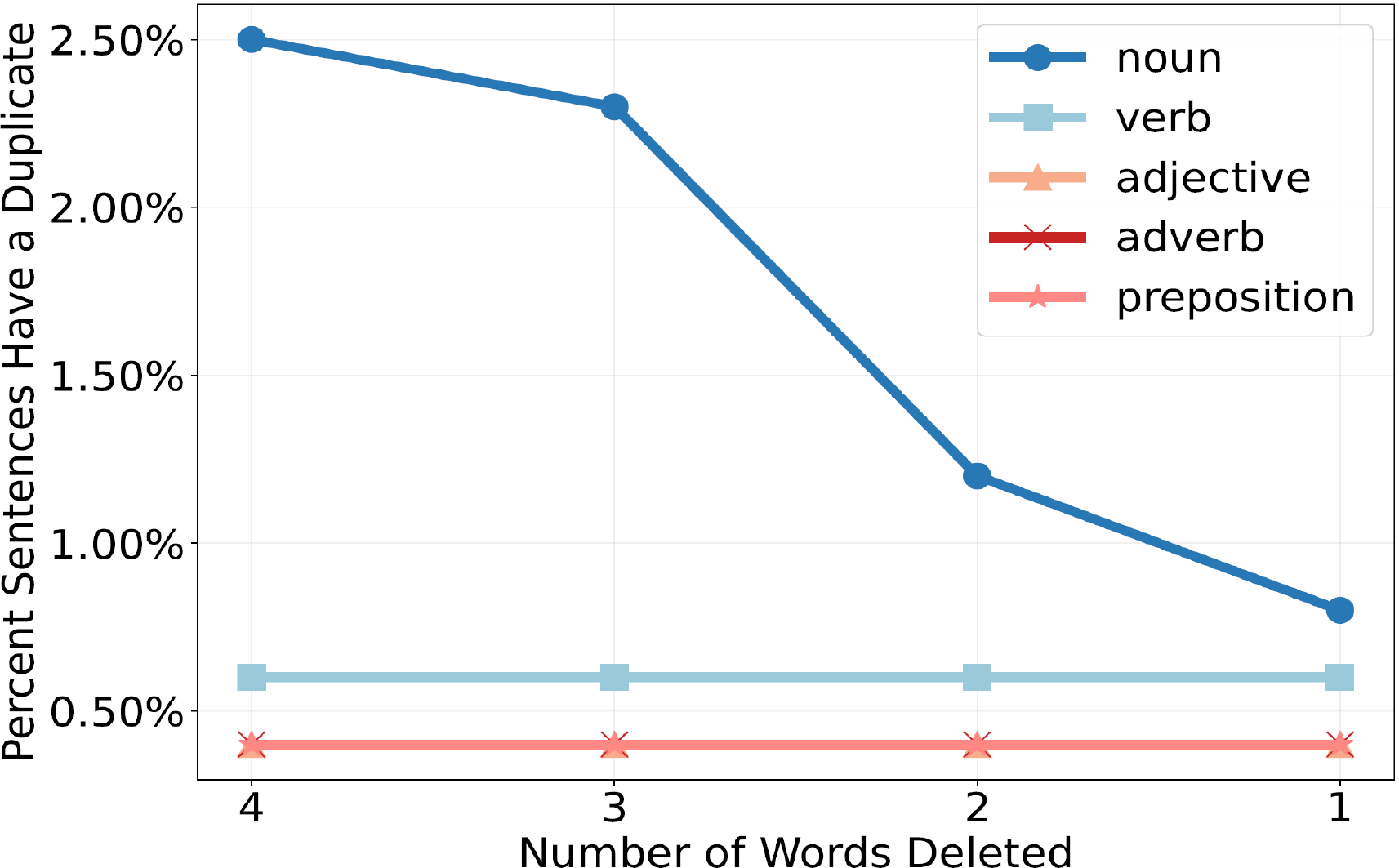}
        \caption{\textbf{MSR-VTT}}
        \label{fig:sub1}
    \end{subfigure}
    
    \begin{subfigure}[b]{\linewidth}
    \vspace{-0.2em}
        \includegraphics[width=0.95\textwidth]{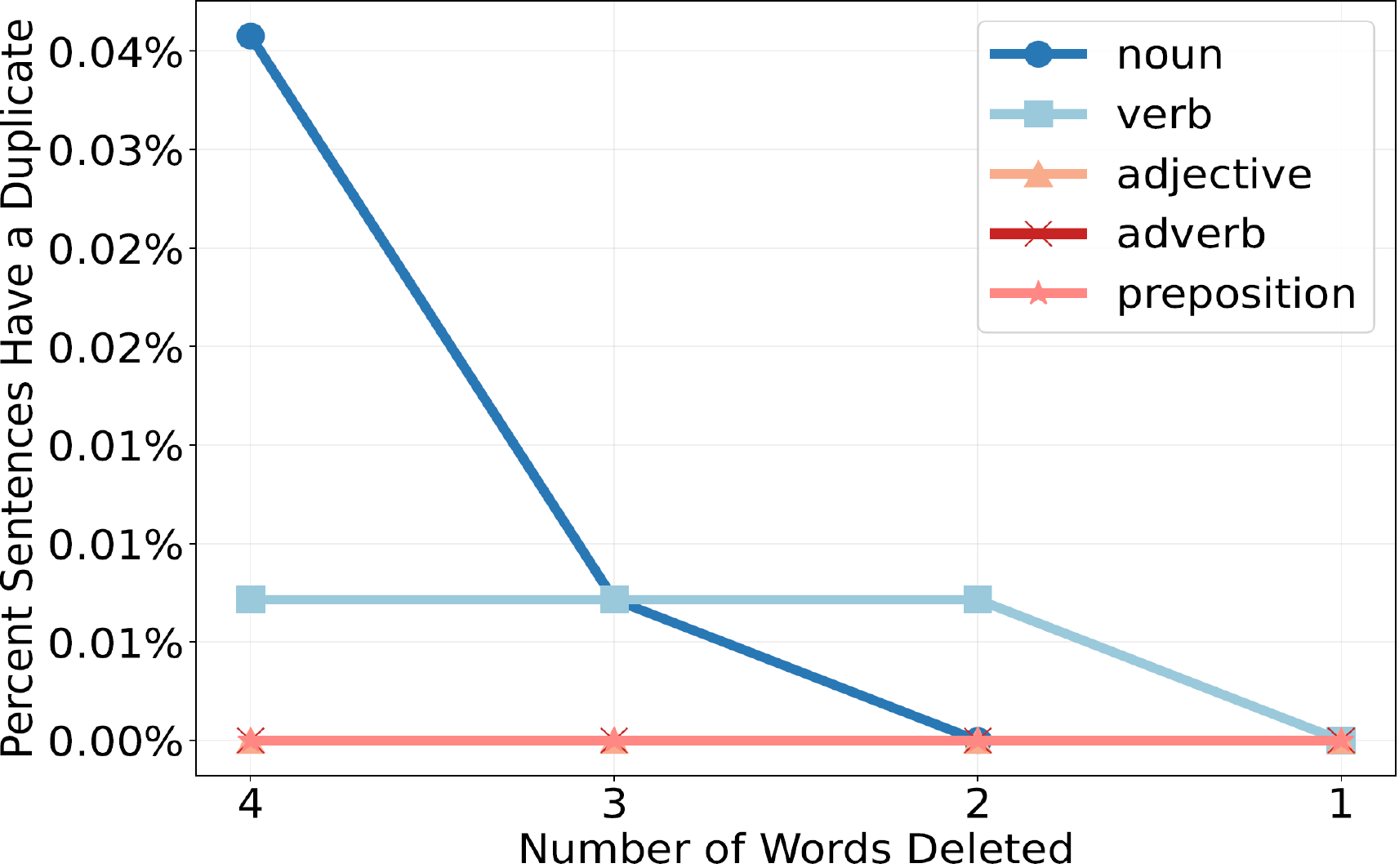}
        \caption{\textbf{VATEX}}
        \label{fig:sub2}
    \end{subfigure}
    \vspace{-1.8em}
    \caption{\textbf{Analysis of Sentence Similarity.} Few captions differ by one part-of-speech, particularly outside of nouns.}
    \vspace{-3em}
     \label{fig:pos-deletion}
\end{wrapfigure}on MSRVTT and VATEX in~\cref{fig:pos-deletion}. Specifically, we randomly
remove a noun from each sentence in the test set and count how many sentences become identical after the removal. We do the same for two nouns, and so forth, until all nouns are removed, thereby testing whether a caption is the same in all other parts-of-speech. We follow the same procedure for verbs, adjectives, adverbs, and prepositions. From~\cref{fig:pos-deletion} we see that captions rarely differ due to a single word. Even when removing 4 nouns, only 0.25\% of captions have a duplicate with the same remaining words. Of course, captions can have the same semantic meaning with different wording, however, the lack of single-word differences in this analysis combined with the qualitative examples highlights the need for fine-grained testing of current video-text retrieval models. 

\vspace{-1em}
\section{PoSRank: A Fine-grained Evaluation Appoach}
\vspace{-0.5em}
\label{sec:fine-metric}
We propose an approach for fine-grained evaluation of video retrieval which complements existing coarse-grained benchmarks. Particularly, our evaluation focuses on changing individual words to create multiple sentences with subtly different meanings. 
If a model truly understands every part of a sentence, then it should be able to detect these subtle differences between sentences and accurately retrieve the sentence most relevant to the video. 
We separate our evaluation by different parts-of-speech, \ie nouns, verbs, adjectives, adverbs, and prepositions, to allow a better understanding of the fine-grained capabilities of the model.

\begin{figure}[tb]
  \centering
  \includegraphics[height=4.5cm]{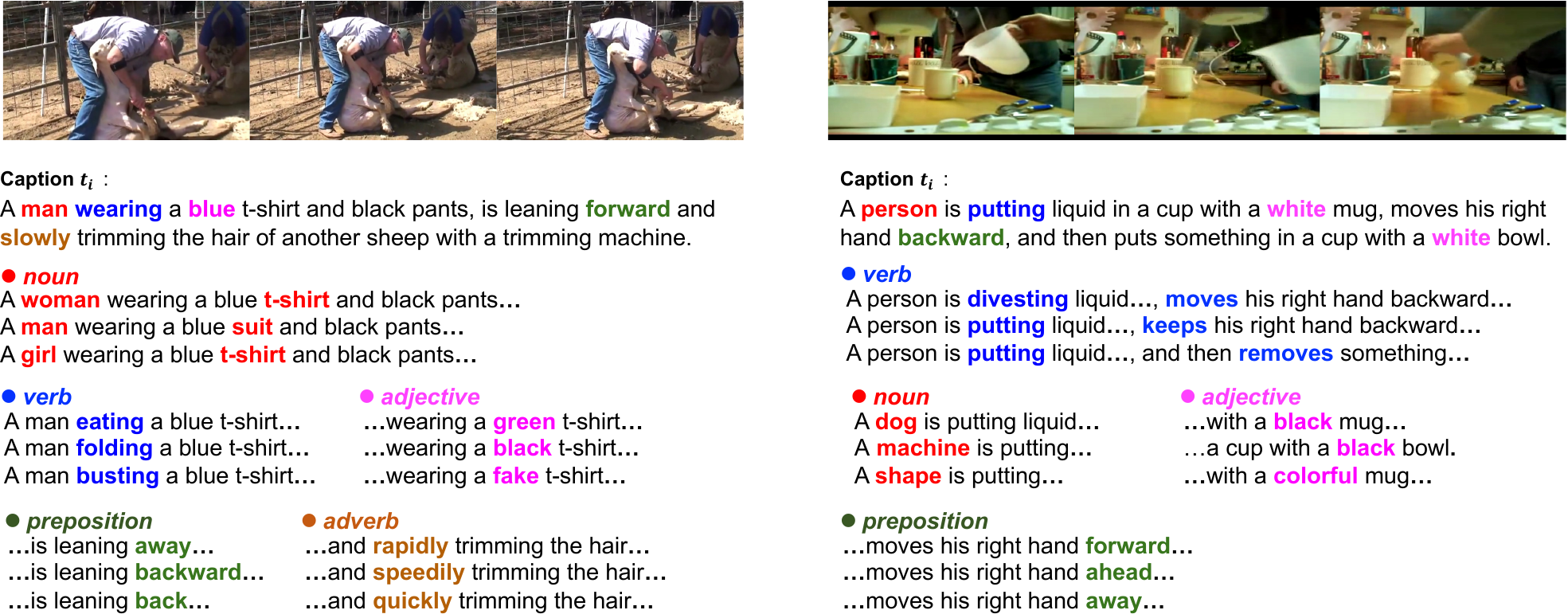}
  \vspace{-0.7em}
  \caption{\textbf{Examples of Hard Negative Sentences} in our evaluation. By creating subtle single-word variations of the original sentence we test the ability of video-text retrieval models to distinguish fine-grained variations for different parts-of-speech.
  }
  \vspace{-1.5em}
  \label{fig:hard_neg}
\end{figure}

\noindent\textbf{Definition. }
We assume we have a set of videos $V$ with a corresponding set of coarse-grained captions $T$.
Given a video $v_i \in V$ and its corresponding caption $t_i \in T$, we create a set of negative captions $T_i^{p}$ of size $N$ for video $v_i$ for each part-of-speech $p\in P$ we are interested in. We create $T_i^{p}$ by
modifying single words with the relevant part-of-speech tag from the caption $t_i$ to generate a fine-grained hard negative sentence. Specifically, for each part-of-speech, we randomly select a word $w$ from sentence $t$ with that part-of-speech tag and modify it to word $w'$ keeping all other words the same. We do this $K$ times to acquire $K$ different sentences which make up set $T_i^{p}$. When selecting word $w'$ we prioritize the antonym of word $w$, followed by the antonym of a hypernym or hyponym of the word. Otherwise, we randomly select a word of the same part-of-speech from the dataset vocabulary. By ensuring the part-of-speech tag is the same we ensure that the sentence is grammatically correct and increase the challenge of the evaluation. Prioritizing antonyms also helps the plausibility of the sentence, increases the challenge and avoids false negatives. From inspecting a random subset of 100 captions we found only 4 false negatives and 7 grammatical errors.
For each video $v_i$ we create negative caption sets for nouns $T_i^{noun}$, verbs $T_i^{verb}$, adjectives $T_i^{adj}$, adverbs $T_i^{adv}$, and prepositions $T_i^{prep}$.  \cref{fig:hard_neg} shows examples of the hard negative sentences created in this process. Only changing a single word per caption results in challenging captions where the model has to distinguish fine-grained concepts \eg `slow' vs. `fast', `forward' vs. `backward' as well as color, age, and gender.



\noindent\textbf{Evaluation Criteria. }
We utilize the newly created negative sentences to evaluate the fine-grained performance of video retrieval models. Specifically, for a part-of-speech $p$ and each video $v_i$, we obtain the ranking of the ground-truth sentence $t_i$ and negatives $T_i^p$.
Suppose the rank of the ground-truth $t_i$ within the retrieved results, including both the correct sentence and the negatives, is denoted by $r^p_i$. We compute the Mean Reciprocal Rank (MRR) for part-of-speech $p$, named $PoSRank^p$, as follows:
\vspace{-1.5em}
\begin{equation}
PoSRank^p = \frac{1}{N} \sum_{i=1}^{N} \frac{1}{r_i^p},
  \label{eq:mrr}
  \vspace{-0.5em}
\end{equation}

where $N$ is the number of ground-truth video-text pairs ($v_i, t_i$).
In summary, $PoSRank^p$ measures the effectiveness of the retrieval model in ranking the ground-truth sentence as close to the top of the list as possible across the negative sets $T_i^p$ of each video. If the retrieval model robustly understands the semantics of each sentence part, the original caption should always rank first. Thus, a higher value indicates a better fine-grained understanding of the model with 1.0 being the best score and 0.0 the worst.


\vspace{-0.8em}
\section{Analysing Fine-grained Ability of Current Models}
\vspace{-0.8em}
\label{sec:fine_eval}
We evaluate existing video-text retrieval methods using our proposed fine-grained evaluation to answer the following questions: (i) How well do current methods understand fine-grained differences in video retrieval? (ii) Does performance in fine-grained retrieval correlate with existing coarse-grained metrics? (iii) Are certain parts-of-speech more challenging than others for current methods? (iv) Are certain datasets more suitable than others for fine-grained evaluation?


\vspace{-1em}
\subsection{Experiment Setup}
\vspace{-0.5em}

\noindent\textbf{Datasets.} 
According to the approach described in \cref{sec:fine-metric}, we create fine-grained evaluation sets for the commonly used MSRVTT and VATEX. Additionally, we use VLN-UVO  and VLN-OOPS \cite{Voigtlaender23vln} originally proposed for video narrative grounding, as these contain longer, more detailed descriptions and thus have a greater number of different part-of-speech with which to create new fine-grained captions, \eg MSR-VTT has 9.3 words per sentence vs. 17.2 from VLN-UVO. 
VLN-UVO and VLN-OOPS also have higher percentages of adjectives, adverbs, and prepositions. For instance, in VLN-OOPS 61\% of sentences contain at least one adverb, while in MSR-VTT only 9\% of sentences have an adverb. 
More detailed statistics can be seen in \cref{tab:createddata-statistics}. 

\begin{table}[t]
\caption{\textbf{Statistics of the fine-grained test sets}. We show the number of original captions, their average length, and the proportions of captions containing each part-of-speech. For all datasets, we create a challenging fine-grained test set per part-of-speech.}
\vspace{-1em}
\label{tab:createddata-statistics}
\centering
\resizebox{\textwidth}{!}{
\begin{tabular}{lccclrrrrrlrrrrr}
\toprule
\multirow{2}{*}{Dataset} & \multicolumn{3}{c}{Original  Captions} &  & \multicolumn{5}{c}{Proportions of Captions} &  & \multicolumn{5}{c}{ Fine-Grained Test Captions} \\
\cmidrule{2-4} \cmidrule{6-10} \cmidrule{12-16} 
 & \multicolumn{1}{c}{\#Train} & \multicolumn{1}{c}{\#Test} & \multicolumn{1}{c}{Av. Len} &  & \multicolumn{1}{c}{noun} & \multicolumn{1}{c}{verb} & \multicolumn{1}{c}{adj} & \multicolumn{1}{c}{adv} & \multicolumn{1}{c}{prep} &  & \multicolumn{1}{c}{noun} & \multicolumn{1}{c}{verb} & \multicolumn{1}{c}{adj} & \multicolumn{1}{c}{adv} & \multicolumn{1}{c}{prep} \\ \midrule
 MSR-VTT \cite{xu2016msr} & 180,000 & 20,000 & 9.3 &  & 1.00 & 0.92 & 0.36 & 0.09 & 0.71  &  & 418.13K & 367.56K & 135.87K & 17.09K & 8.51K  \\
VATEX \cite{wang2019vatex} & 519,820 & 13,969 & 14.3 &  & 1.00 & 0.99 & 0.49 & 0.24 & 0.88 &  & 293.35K & 283.40K & 134.59K & 32.38K & 12.12K \\
VLN-UVO \cite{Voigtlaender23vln} & 12,359 & 6,561 & 17.2 &  & 1.00 & 1.00 & 0.89 & 0.26 & 0.86 &  & 137.74K & 136.48K & 120.23K & 4.58K & 4.96K  \\
VLN-OOPS \cite{Voigtlaender23vln} & 26,193 & 5,335 & 25.7 &  & 1.00 & 1.00 & 0.92 & 0.61 & 0.98 &  & 112.04K & 111.62K & 99.31K & 18.61K & 6.38K  \\ \bottomrule
\end{tabular}
}
\vspace{-1em}
\end{table}

\noindent\textbf{Baselines.} 
We use four recent, publicly available video retrieval models covering a variety of different backbones and training strategies. 
\textbf{Frozen} \cite{bain2021frozen} is a joint embedding model for end-to-end video retrieval.  
It utilizes a ViT backbone pre-trained on CC3M\cite{sharma2018gcc} (image-text pairs) and WebVid-2M\cite{bain2021frozen} (video-text pairs)  with a contrastive loss. 
\textbf{TS2Net} \cite{ts2net} initializes its visual and text encoders with CLIP. It uses a token shift module which aims to capture subtle local movements between frames as well as a token selection module to minimize spatial and temporal redundancy.
%
\textbf{X-CLIP} \cite{xclip} also bases its model on CLIP and focuses on computing fine-grained cross-modal similarities. Specifically, it obtains the final video-text similarity by fusing video-sentence, video-word, sentence-frame, and frame-word similarities. 
\textbf{UCoFiA} \cite{wang2023ucofia} aligns text and video at both coarse and fine-grained granularities. This is achieved by three types of matching: video-sentence, frame-sentence, and patch-word. This model is also based on CLIP. 


\noindent\textbf{Implementation Details. }
For each original caption, we create $K{=}20$ negative captions for each of the five parts-of-speech using SpaCy for parsing. 

\vspace{-1.0em}
\subsection{Results}
\vspace{-0.5em}

\noindent\textbf{How well do current methods understand fine-grained differences?}
We show the performance of the four tested video-text retrieval models with our \textit{PoSRank} metric, averaged over the five parts-of-speech, 
in \cref{tab:av_pos}.  For all methods and datasets, the \textit{PoSRank} scores range from $0.24$ to $0.36$ far from the best score of 1. This demonstrates that current methods have poor fine-grained understanding and are not sensitive to subtle single-word changes in the caption. CLIP-based methods X-CLIP, UCoFiA and TS2-Net all perform slightly better than Frozen. This may be due to the CLIP-based methods having a better understanding of different parts-of-speech or due to these approaches learning better video-text representations. We analyze both possibilities in the following questions. Regardless, all methods evaluated struggle with our proposed fine-grained evaluation and have much room for improvement.



\begin{table}[t]
\caption{\textbf{Fine-Grained Evaluation with Average PoSRank}. All four methods struggle to distinguish fine-grained differences in all datasets.}
\vspace{-0.8em}
\label{tab:av_pos}
\centering
\resizebox{0.65\textwidth}{!}{
\begin{tabular}{lcccccc}
\toprule
Method & & \multicolumn{1}{l}{MSR-VTT} & \multicolumn{1}{l}{VATEX} & \multicolumn{1}{l}{VLN-UVO} & \multicolumn{1}{l}{VLN-OOPS}  & \multicolumn{1}{l}{Mean} \\ \midrule
Frozen \cite{bain2021frozen} & & 0.285 & 0.243 & 0.252 & 0.249 & 0.257 \\
X-CLIP \cite{xclip} &  & 0.343 & 0.278 & 0.301 & 0.282 & 0.301 \\
UCoFiA \cite{wang2023ucofia} & & 0.351 & 0.268 & 0.308 & 0.299 & 0.306 \\
TS2-Net \cite{ts2net} & & 0.351 & 0.283 & 0.310 &0.293 &0.309\\

\bottomrule
\end{tabular}}
\vspace{-0.5em}
\end{table}

\begin{figure}[t]
    \centering
   
    \begin{subfigure}[b]{0.24\textwidth}
        \includegraphics[width=\textwidth]{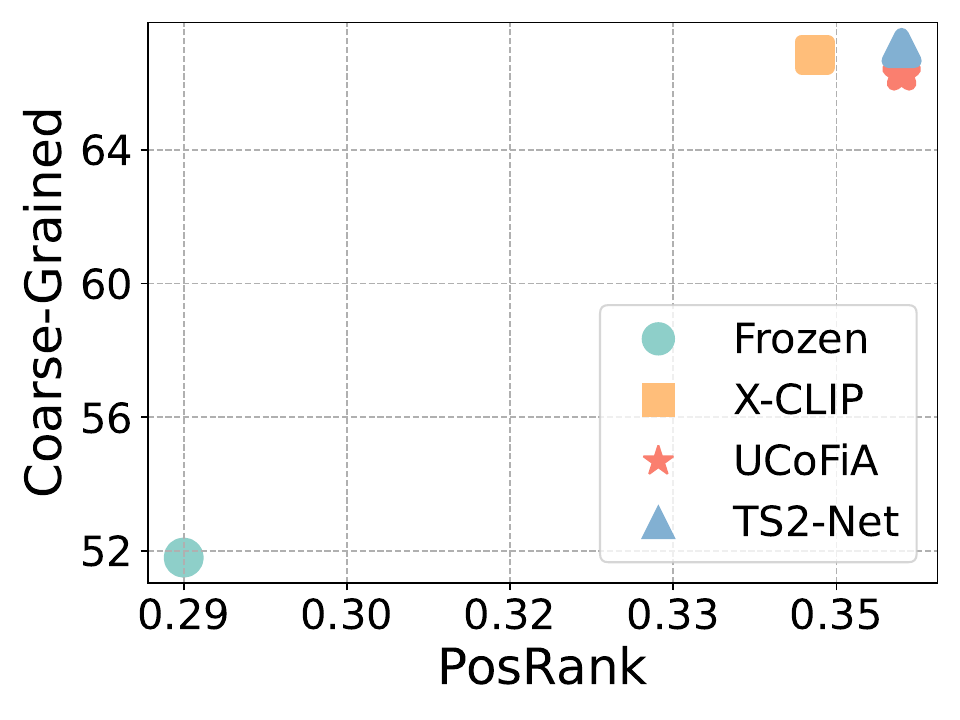}
        \caption{\textbf{MSR-VTT}}
        \label{fig:sub1}
    \end{subfigure}
    \begin{subfigure}[b]{0.24\textwidth}
        \includegraphics[width=\textwidth]{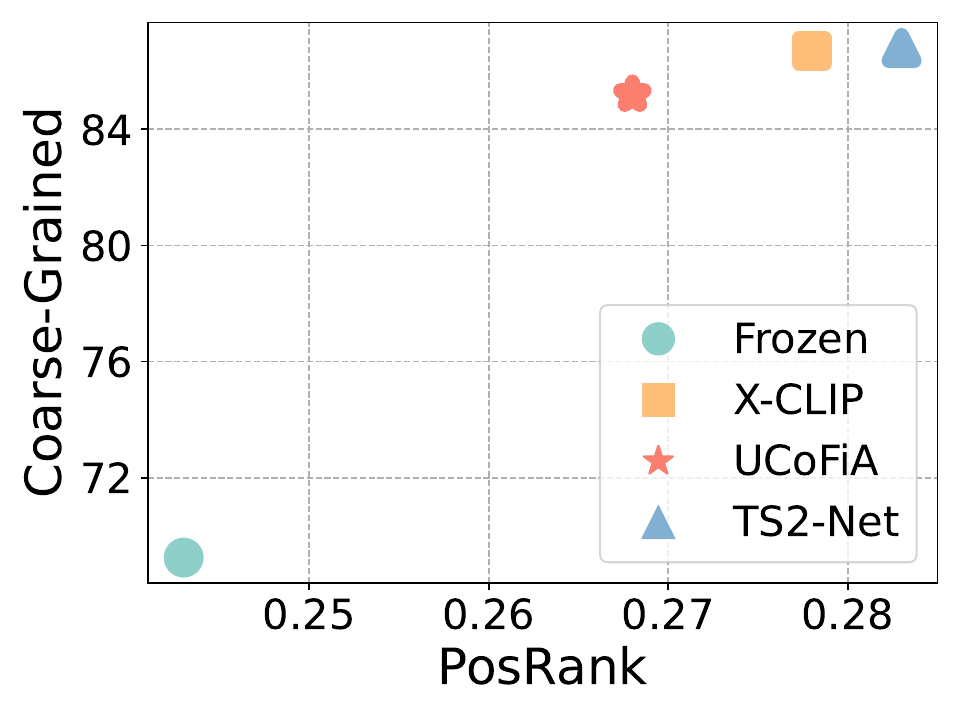}
        \caption{\textbf{VATEX}}
        \label{fig:sub2}
    \end{subfigure}
    \begin{subfigure}[b]{0.24\textwidth}
        \includegraphics[width=\textwidth]{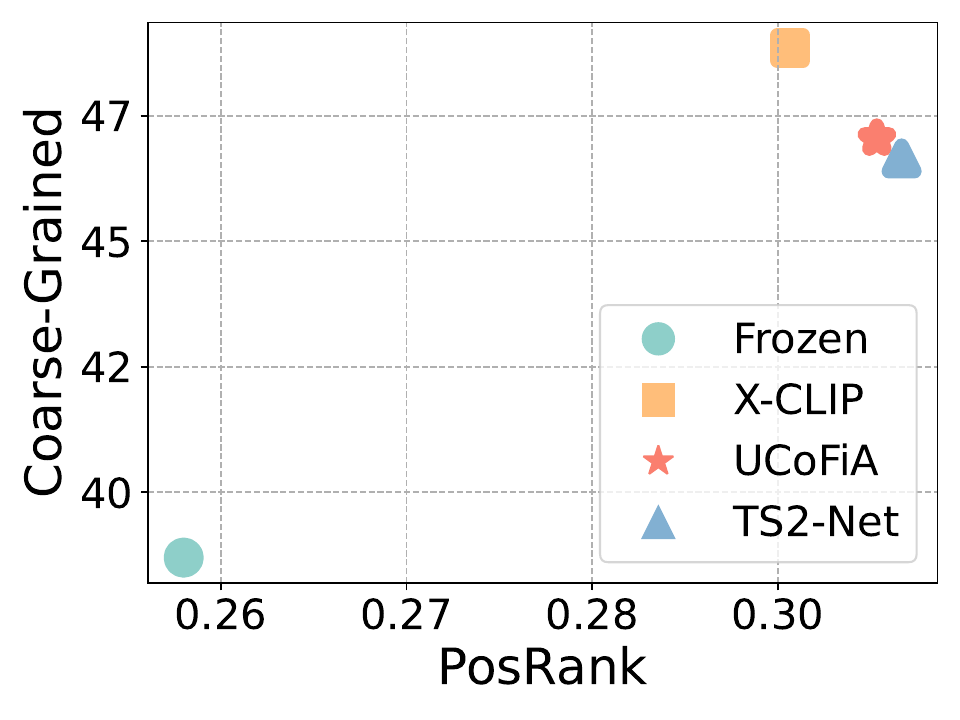}
        \caption{\textbf{VLN-UVO}}
        \label{fig:sub3}
    \end{subfigure}
    \begin{subfigure}[b]{0.24\textwidth}
        \includegraphics[width=\textwidth]{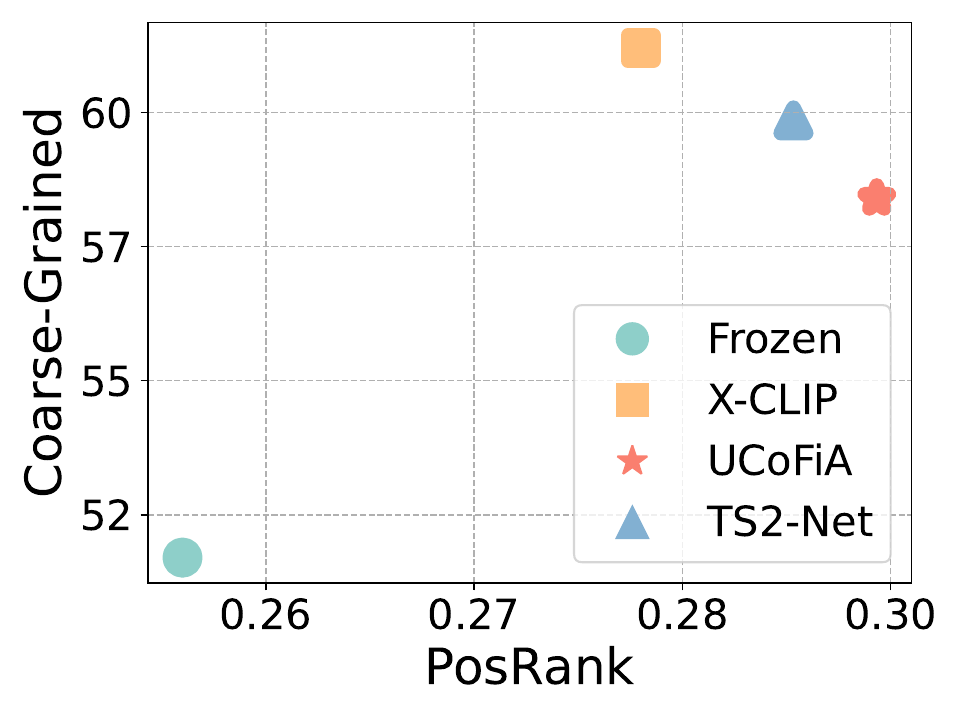}
        \caption{\textbf{VLN-OOPS}}
        \label{fig:sub4}
    \end{subfigure}
\vspace{-0.5em}
    \caption{\textbf{Correlation Between Coarse \& Fine-Grained Performance.}  There is a weak correlation between performance on coarse and fine-grained metrics, however good coarse-grained performance does not indicate good fine-grained performance.}
     \label{fig:coarse_fine+corr}
     \vspace{-1em}
\end{figure}

\noindent\textbf{Does performance in fine-grained retrieval correlate with existing coarse-grained metrics?}
\cref{fig:coarse_fine+corr} shows mean recall@\{1,5,10\} of video-to-text and text-to-video retrieval alongside the mean \textit{PoSRank}. There is some correlation between performance on coarse and fine-grained metrics. 
For instance, Frozen is the worst in both coarse-grained and fine-grained retrieval. However, at the higher end the correlation is weak with the best-performing model in coarse-grained achieving only 3rd place on fine-grained for VLN-UVO and VLN-OOPS.
This demonstrates that good coarse-grained performance does not indicate good fine-grained performance and further justifies the need for our fine-grained \textit{PoSRank} evaluation.
Furthermore, the differences between methods in coarse-grained metrics can be large, \eg $71$ for Frozen vs. $85$ for UCoFiA on VATEX, while performance is more similar in our fine-grained \textit{PoSRank}, \eg $0.24$ vs. $0.27$. 
This demonstrates that all current methods, regardless of their success on current coarse-grained metrics, have a long way to go in fine-grained understanding. 

\begin{figure}[t]
    \centering
   
    \begin{subfigure}[b]{0.35\textwidth}
        \includegraphics[width=\textwidth]{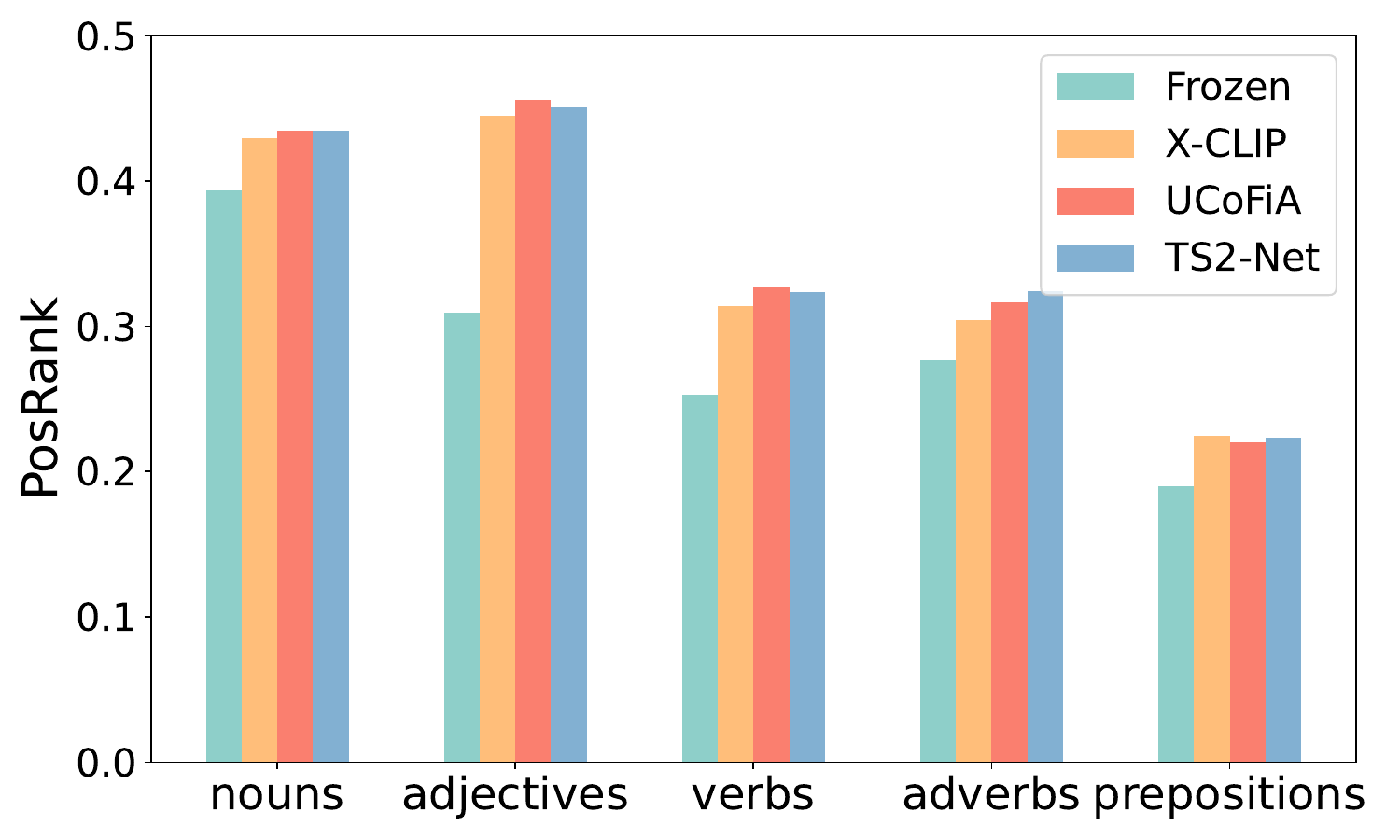}
        \caption{\textbf{MSR-VTT}}
        \label{fig:sub1}
    \end{subfigure}
    \hspace{4mm} 
     \hspace{4mm} 
    \begin{subfigure}[b]{0.35\textwidth}
        \includegraphics[width=\textwidth]{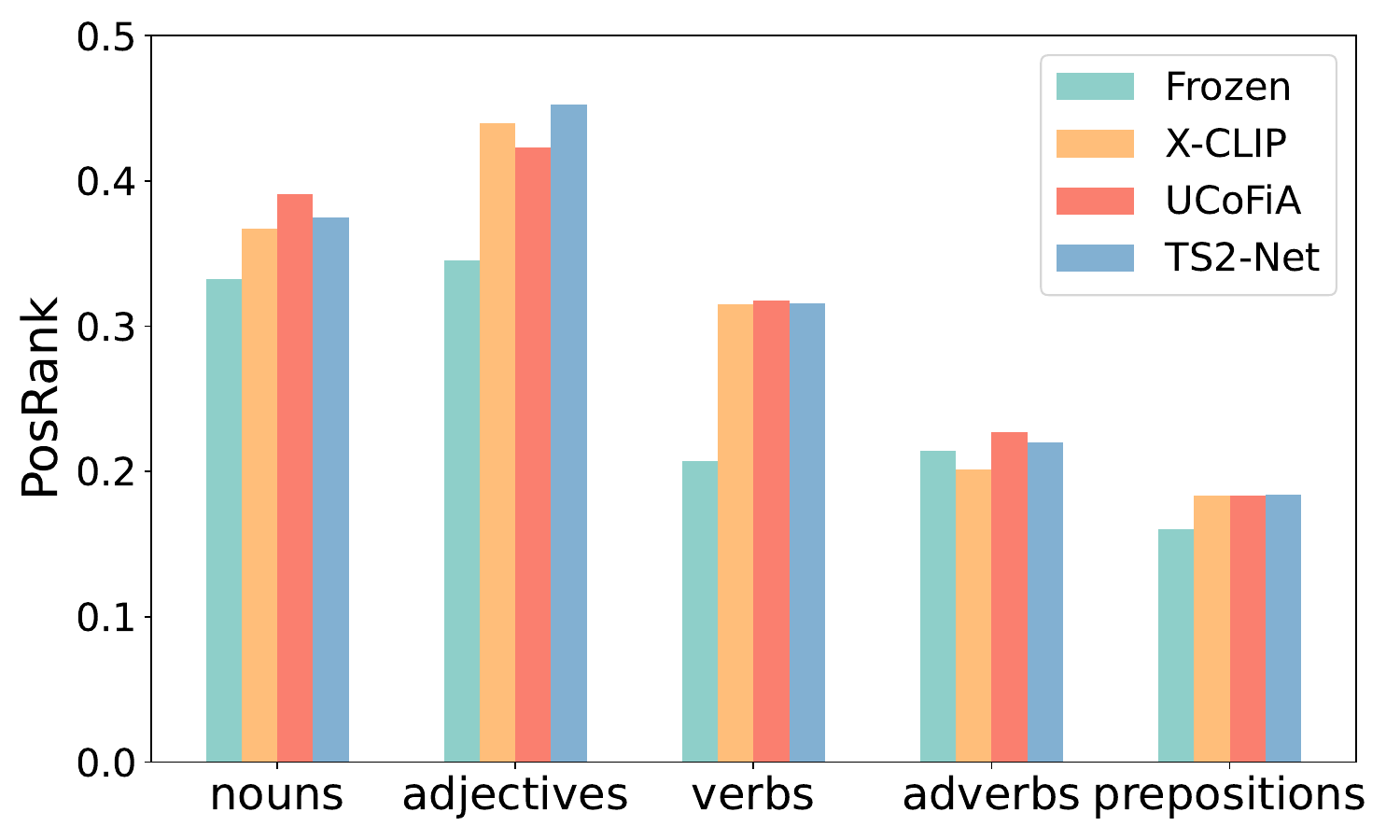}
        \caption{\textbf{VLN-UVO}}
        \label{fig:sub3}
    \end{subfigure}
\vspace{-0.5em}
    \caption{\textbf{Fine-Grained Evaluation Per Part-of-Speech.} Models find fine-grained differences in adverbs and prepositions the most difficult to distinguish. }
    \vspace{-0.8em}
     \label{fig:fine_eval_bar_baseline_data}
\end{figure}

\noindent\textbf{Are certain parts-of-speech more challenging than others?}
\cref{fig:fine_eval_bar_baseline_data} shows \textit{PoSRank} performance for individual parts-of-speech for all methods on MSR-VTT and VLN-UVO. 
From these results, we see that current models can better distinguish differences in adjectives, with nouns being the second easiest. This is likely due to the reliance of video-text retrieval models on image pretraining where nouns and adjectives are common; three of the four methods make use of CLIP models while Frozen also pretrains on images and single video frames. Adverbs and prepositions are the hardest parts-of-speech on both datasets for almost all models. 
This is likely due to these parts-of-speech requiring more understanding of the spatial and temporal relationships between different objects and frames, unlike nouns and adjectives which can often be understood from the appearance of a single object in a single frame. Results for VATEX and VLN-OOPS follow similar trends and can be found in the supplementary material.

\noindent\textbf{Are certain datasets more suitable than others for fine-grained evaluation?}
The fine-grained sets we create for VATEX, VLN-UVO and VLN-OOPS are all more challenging for current methods than the one created for MSR-VTT. 
For instance, the average \textit{PoSRank} value for XCLIP is $0.278$ on VATEX and $0.282$ on VLN-OOPS, while on MSR-VTT, it is $0.343$ (\cref{tab:av_pos}). This is due to longer and more complex original captions offering more potential for challenging and subtle variations that better test the fine-grained capabilities of a model. VLN-UVO and VLN-OOPS also provide more challenges for the most difficult parts-of-speech. For instance, on VLN-UVO models achieves $\sim0.2$ adverb \textit{PoSRank}, compared to $\sim0.3$ on MSR-VTT (\cref{fig:fine_eval_bar_baseline_data}).  

\vspace{-1em}
\section{Training for Fine-grained Understanding}
\begin{figure}[tb]
  \centering
  \includegraphics[width=0.8\linewidth]{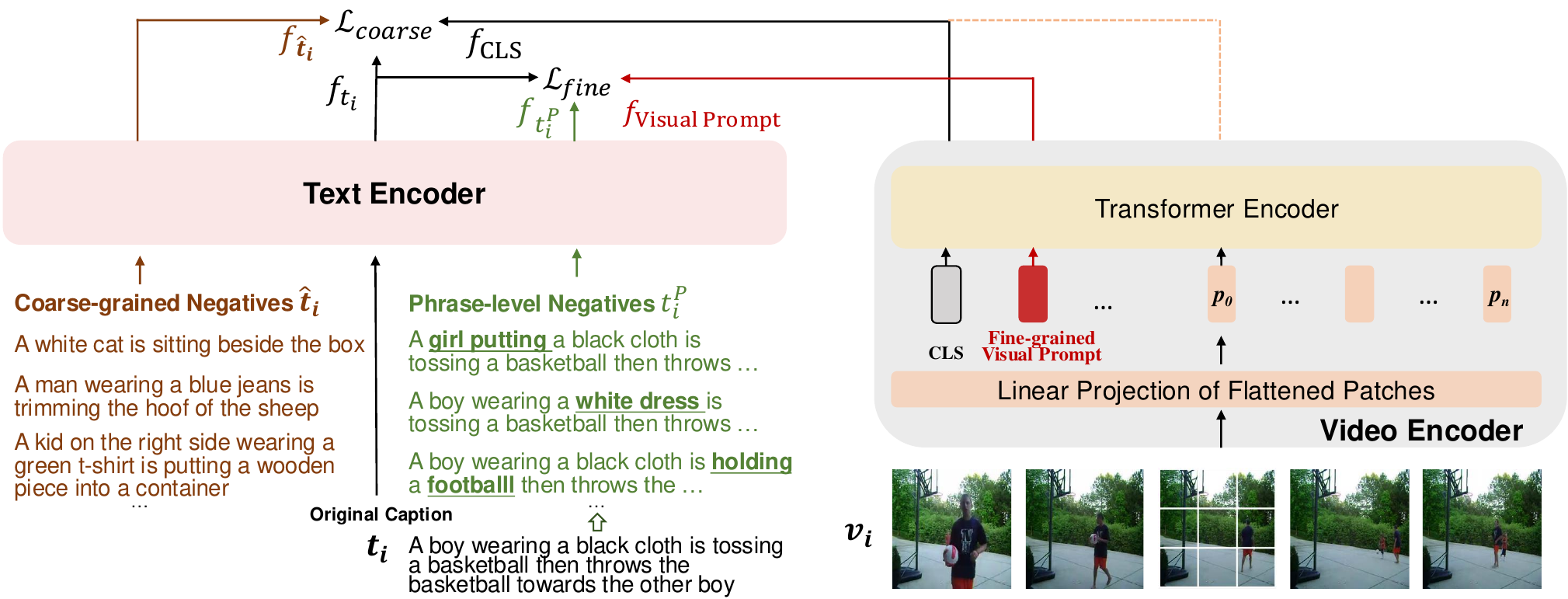}
  \vspace{-0.5em}
  \caption{\textbf{Fine-grained Training for Video-Text Retrieval} incorporates phrase-level negatives and fine-grained prompting to identify coarse and fine-grained differences. 
  \vspace{-1.5em}
  }
  \label{fig:method}
\end{figure}

\vspace{-0.5em}
Our fine-grained evaluation shows there is still much room for improvement in current methods. To improve fine-grained understanding, we introduce a new baseline (\cref{fig:method}) that can be easily combined with existing approaches. Beginning from an existing coarse-grained video-text retrieval model (\cref{sec:coarse_training}), we train with fine-grained `word-level' negatives (\cref{sec:fine-grained-traning}). While this improves the model's ability to distinguish fine-grained text differences, it hurts the ability to distinguish coarse-grained differences. We strike a balance between coarse and fine with `phrase-level negatives' (\cref{sec:phrase}) and better allow the model to lean from both types of negatives with fine-grained prompts (\cref{sec:vpt}).

\vspace{-1em}
\subsection{Coarse-Grained Training}
\vspace{-0.5em}
\label{sec:coarse_training}
A typical retrieval model learns from a set of videos $V$ and a corresponding set of coarse-grained captions $T$. As in coarse-grained evaluation each caption $t_i\in T$ is paired with a single video $v_i \in V$, although a video may have multiple corresponding captions. The goal is to give matching video text pairs $(v_i, t_i)$ high similarity. This similarity should be higher than the same video with a less relevant text $(v_i, t_i')$ or the same text with a less relevant video $(v_i', t_i)$. Assuming all unpaired videos and texts are less relevant to each other than the pairs, a model can be trained with the following $\mathcal{L}_{coarse}$ loss:
\vspace{-0.5em}
\begin{equation}
\begin{aligned}
\label{eq:coarse}
&\mathcal{L}_{v2t} =-\frac1B\sum_{i=1}^B\log\frac{\exp\left(s(v_i,t_i)\right)}{\sum_{j=1}^B\exp\left(s(v_i,t_j)\right)}, \\
&\mathcal{L}_{t2v} =-\frac1B\sum_{i=1}^B\log\frac{\exp\left(s(v_i,t_i)\right)}{\sum_{j=1}^B\exp\left(s(v_j,t_i)\right)}, \\
&\mathcal{L}_{coarse}=\mathcal{L}_{v2t}+\mathcal{L}_{t2v},
\end{aligned}
\end{equation}
where $s(\cdot)$ is a similarity function and $B$ is the batch size. 
While this loss is effective for training models for existing coarse-grained video retrieval metrics, models trained in this way do not learn to distinguish fine-grained word-level differences. This is because just like the test set (see ~\cref{sec:coarse-metric}) the set of captions $T$ in training does not contain fine-grained differences. 

\vspace{-1em}
\subsection{Training with Word-Level Negatives}
\vspace{-0.5em}
\label{sec:fine-grained-traning}
In addition to using our sets of fine-grained negative captions $T^{p}_i$ for evaluation, we take inspiration from prior works on hard negative\cite{a-neurips20-hardnegmix, b-iclr21-contrastive} mining and also incorporate our fine-grained negatives during training to enhance the model's sensitivity to all parts of a sentence. 
For each video-text pair $(v_i, t_i)$ in a batch, we generate $\mathbb{N}$ word-level negative sentences for each of the five parts-of-speech we are interested in, using the method described in Section \cref{sec:fine-metric}. 
We incorporate these word-level negatives into training using the following contrastive loss:
\vspace{-0.8em}
\begin{equation}
\label{eq:fine}
\begin{aligned}
&\mathcal{L}_{fine}=-\frac1B\sum_{i=1}^B\log\frac{\exp\left(s(v_i,t_i)\right)}{\sum_{p\in P}\sum_{t \in T_i^p} \exp\left(s(v_i,t)\right)}, 
 \end{aligned}
 \vspace{-0.3em}
\end{equation}
where $P$ is the set of POS-tags of interest and $T_i^p$ is the set of word-level negative captions created from caption $t_i$ for POS-tag $p$.

Since we want the model to be both suitable for coarse and fine-grained evaluation we combine the two losses as follows:
\vspace{-0.4em}
\begin{equation}
\label{eq:overall}
\begin{aligned}
&\mathcal{L}=\mathcal{L}_{coarse}+\lambda \mathcal{L}_{fine},
 \end{aligned}
 \vspace{-0.4em}
\end{equation}
where $\lambda$ is the weight to balance the coarse and fine loss. We use $\lambda \ll 1$ as the similarity between a video and its word-level hard negatives should be greater than the similarity between a video and the coarse-grained negatives. 

\vspace{-1em}
\subsection{Training with Phrase-Level Negatives}
\vspace{-0.5em}
\label{sec:phrase}
 Training with the loss in ~\cref{eq:overall} enables the model to make fine-grained word-level distinctions, particularly for nouns, adjectives and verbs which are plentiful in the training data enabling the creation of a large variety of negatives. However, even with this loss models struggle to balance coarse and fine-grained, with results decreasing on coarse-grained metrics as fine-grained performance increases. 
 Due to the vastly different objectives of coarse and fine-grained video retrieval current models struggle to learn both tasks simultaneously. 

To address this we propose using `phrase-level' negatives which balance coarse and fine granularities. Our phrase-level negatives modify two consecutive words in a sentence.  As with the word-level negatives, for a video $v_i$ with corresponding caption $t_i$ we randomly select a word $w_1$ from caption $t_i$ to modify which is one of the five parts-of-speech of interest. Additionally, with 50:50 chance we either modify the preceding or following word which is also one of the five parts-of-speech. 
For example, if \textit{black} is the word selected from \textit{a boy wearing a black t-shirt} then we could obtain the following phrase-level negatives: \textit{a boy removing a white t-shirt} or \textit{a boy wearing a black coat}. 
%
As in the word-level negatives and fine-grained evaluation, words are replaced with other words from the dataset with the same POS-tag, prioritizing antonyms. All remaining words are unmodified.

\vspace{-1em}
\subsection{Fine-Grained Prompting}
\vspace{-0.5em}
\label{sec:vpt}
Our phrase-level negatives allow the model to better balance between coarse and fine-grained distinctions. However, models still struggle to optimize for both objectives and underperform a model specialized for either coarse or fine-grained. To remedy this we incorporate prompts to inform the model which task it should focus on. 
Specifically, we use visual prompts~\cite{jia2022vpt} to learn fine-grained similarity.
This gives us additional output tokens of the transformer so the same model can predict both fine and coarse-grained similarity.
We train the output of the CLS token with the coarse-grained loss $\mathcal{L}_{coarse}$ (\cref{eq:coarse}), which establishes a general understanding of the content. 
Concurrently, we train the output of the visual prompt with $\mathcal{L}_{fine}$ (\cref{eq:fine}), enhancing the model's capability to detect subtle nuances. 
This allows for a comprehensive and detailed understanding of video content, aligning video with the text queries more precisely.

\vspace{-0.5em}
\section{Experiments}
\vspace{-1.0em}
\label{sec:experiment}
We first describe the implementation details and then analyze the contribution of each method component and the impact of introduced hyperparameters. Finally, we demonstrate that our method can be plugged into various existing video-text retrieval models and is effective on multiple different datasets. 

\noindent\textbf{Implementation Details. }
We use the same datasets and methods as described in ~\cref{sec:fine_eval}. For all models we use the same training settings as the original paper with the addition of our fine-grained training, see the supplementary material for further details.
Unless otherwise specified, we use loss weight $\lambda{=}0.2$, number of hard negatives $\mathbb{N}{=}16$, and number of visual prompts $\mathbb{T}{=}1$. All experiments were performed on an NVIDIA RTX A6000 GPU.

\noindent\textbf{Evaluation Criteria.} For coarse-grained evaluation, we report average  recall over R@1, R@5 and R@10. For fine-grained evaluation, we report the proposed \textit{PosRank} for nouns, adjectives, verbs, adverbs and prepositions.

\vspace{-1em}
\subsection{Ablation Study}
\vspace{-1em}
To validate the effectiveness our proposed method, we first carry out several ablation studies using the X-CLIP model on the VLN-UVO dataset. 




\begin{table}[t]
\caption{\textbf{Model Components Ablation}. Phrase-level negative with fine-grained prompting balances  coarse (mean recall) and fine-grained (PoSRank) understanding.} 

\vspace{-1em}
\label{tab:component_ablation}
\centering
\setlength{\tabcolsep}{2pt}
\resizebox{0.7\linewidth}{!}{
\begin{tabular}{lw{c}{4em}w{c}{4em}cccccc}
\toprule
\multirow{2}{*}{Training-strategy}  & \multicolumn{2}{c}{Coarse-Grained ($\uparrow$)} &  & \multicolumn{5}{c}{Fine-Grained ($\uparrow$)} \\
\cmidrule{2-3} \cmidrule{5-9}
 & V2T & T2V & & \multirow{1}{*}{noun} & \multirow{1}{*}{adj} & \multirow{1}{*}{verb} & \multirow{1}{*}{adv} & \multirow{1}{*}{prep} \\
 \midrule
Coarse-Grained Training & 56.7 & 41.0 &  & 0.367 & 0.440 & 0.315 & 0.201 & 0.184 \\
Word-Level Negatives & 47.5 & 40.0 &  & 0.894 & 0.864 & 0.969 & 0.468 & 0.701 \\
Phrase-Level Negatives & 52.5 & 40.2 &  & 0.839 & 0.723 & 0.913 & 0.382 & 0.527 \\
Fine-Grained Prompting & 52.9 & 39.8 &  & 0.860 & 0.781 & 0.954 & 0.439 & 0.555 \\
 \bottomrule
\end{tabular}}
\vspace{-1em}
\end{table}

\begin{figure}[t!]
  \centering
  \includegraphics[width=\linewidth]{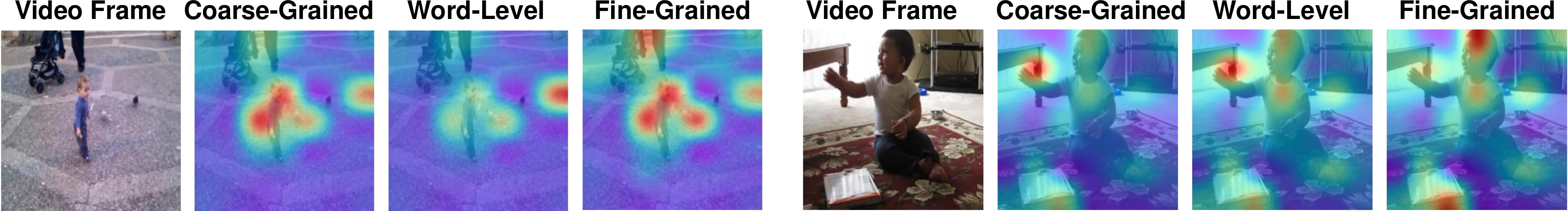}
  \vspace{-2.2em}
  \caption{\textbf{Visual Attention.} Coarse-grained training focuses on the most salient regions, while word-level negatives focuses on subtle details necessary for distinguishing fine-grained elements of the caption. Fine-grained prompting balances the two.
   \vspace{-2em}
  }
  \label{fig:atten-amp}
\end{figure}

\noindent\textbf{Model Components.} ~\cref{tab:component_ablation} demonstrates the effect of each component of our proposed method. Initially, we train X-CLIP~\cite{xclip} as in the original paper, \ie coarse-grained training. This performs well on the standard coarse-grained evaluation metrics but struggles on our fine-grained evaluation, particularly for adverbs and prepositions. When adding fine-grained training with word-level negatives the fine-grained evaluation improves for all parts-of-speech, but this comes at the cost of decreased video-to-text coarse-grained retrieval. This is because the focus is shifted from the most salient regions to fine-grained details (see Figure~\ref{fig:atten-amp}). Replacing the word-level negatives with phrase-level negatives better balances the coarse and fine-grained objectives, for instance, coarse-grained video-to-text retrieval (V2T) increases from 47.5 to 52.5. Using fine-grained prompting with phrase-level negatives further improves the fine-grained evaluation results as it allows the model to separately specialize in coarse- and fine-grained differences.

\noindent
\begin{minipage}[t]{.48\textwidth}
\centering
\vspace{-1em}
\captionof{table}{\textbf{Loss weight $\lambda$}. $\lambda{=}0.2$ is the best balance between fine-grained (PoSRank) and coarse-grained (mean recall).}
\label{tab:ablation2}
\setlength{\tabcolsep}{6pt}
\resizebox{\textwidth}{!}{%
\begin{tabular}{lw{c}{3.2em}w{c}{3.2em}cccccc}
\toprule
\multirow{2}{*}{$\lambda$}  & \multicolumn{2}{c}{Coarse-Grained ($\uparrow$)} &  & \multicolumn{5}{c}{Fine-Grained ($\uparrow$)} \\
\cmidrule{2-3} \cmidrule{5-9}
 & \multirow{1}{*}{V2T}  &  \multirow{1}{*}{T2V} & & \multirow{1}{*}{noun} & \multirow{1}{*}{adj} & \multirow{1}{*}{verb} & \multirow{1}{*}{adv} & \multirow{1}{*}{prep} \\
\midrule
0.1 & 48.2 & 39.8 &  & 0.871 & 0.814 & 0.959 & 0.367 & 0.627 \\
0.2 & 47.5 & 40.0 &  & 0.894 & 0.864 & 0.969 & 0.468 & 0.701 \\
0.3 & 42.7 & 40.0 &  & 0.903 & 0.875 & 0.976 & 0.459 & 0.711 \\
0.4 & 41.4 & 39.5 &  & 0.906 & 0.880 & 0.977 & 0.482 & 0.730 \\
0.5 & 43.3 & 39.4 &  & 0.906 & 0.881 & 0.976 & 0.523 & 0.728\\
\bottomrule
\end{tabular}}
\vspace{-0.5em}
\end{minipage}%
\hfill
\begin{minipage}[t]{.48\textwidth}
\centering
\vspace{-1em}
\captionof{table}{\textbf{Num. of hard negatives $\mathbb{N}$}. $\mathbb{N}=16$ provides a reasonable balance between performance and training efficiency.}
\label{tab:ablation1}
\setlength{\tabcolsep}{6pt}
\resizebox{\textwidth}{!}{%
\begin{tabular}{lw{c}{3.2em}w{c}{3.2em}cccccc}
\toprule
\multirow{2}{*}{$\mathbb{N}$}  & \multicolumn{2}{c}{Coarse-Grained ($\uparrow$)} &  & \multicolumn{5}{c}{Fine-Grained ($\uparrow$)} \\
\cmidrule{2-3} \cmidrule{5-9}
 & \multirow{1}{*}{V2T}  &  \multirow{1}{*}{T2V} & & \multirow{1}{*}{noun} & \multirow{1}{*}{adj} & \multirow{1}{*}{verb} & \multirow{1}{*}{adv} & \multirow{1}{*}{prep} \\
\midrule
2 & 53.3 & 39.9 &  & 0.778 & 0.692 & 0.882 & 0.282 & 0.475 \\
4 & 48.6 & 39.8 &  & 0.875 & 0.808 & 0.957 & 0.344 & 0.655 \\
8 & 48.7 & 39.8 &  & 0.876 & 0.834 & 0.962 & 0.420 & 0.656 \\
16 & 47.5 & 40.0 &  & 0.894 & 0.864 & 0.969 & 0.468 & 0.701 \\
 32 & 44.0 & 40.0 & &  0.901 & 0.869 & 0.975 & 0.449 & 0.701 \\
 \bottomrule 
\end{tabular}} 
\vspace{-0.5em}
\end{minipage}
\vspace{1.5em}

\noindent\textbf{Loss weighting $\lambda$.}
We evaluate the impact of $\lambda$, which balances coarse and fine-grained loss functions, in ~\cref{tab:ablation2}. As $\lambda$ increases, the model focuses on the fine-grained loss, leading to improved fine-grained performance. $\lambda = 0.2$ represents the best balance between fine-grained and coarse-grained performance.

\noindent\textbf{Number of hard negative sentences $\mathbb{N}$.} We explore the number of hard negative sentences $\mathbb{N}$ in~\cref{tab:ablation1}. As $\mathbb{N}$ increases fine-grained retrieval improves but coarse-grained performance declines. 
$\mathbb{N}${=}16 represents a compromise between fine-grained and coarse-grained performance.

\vspace{-1.0em}
\subsection{Effectiveness with Different Datasets.}
\vspace{-0.5em}
We demonstrate our model's effectiveness for different datasets with X-CLIP in ~\cref{tab:sec7_xclip}. On all datasets, our approach provides much better fine-grained retrieval performance than standard coarse-grained training, for instance on VLN-OOPS we obtain 0.941 verb and 0.527 preposition \textit{PoSRank} vs. 0.236 and 0.142 for coarse-grained training. Our approach is also more effective at coarse-grained metrics than the naive fine-grained training with word-level negatives, \eg we obtain 67.8 V2T on VLN-OOPS compared to 62.9 for word-level negatives. However, there is still much potential for improvement in both coarse and fine-grained metrics, particularly for adverbs and prepositions in the fine-grained evaluation.

\begin{table}[htbp]
\caption{\textbf{Effectiveness with Different Datasets}. Our approach recognizes fine-grained differences (PoSRank) on different datasets while  maintaining coarse-grained ability (mean recall). }
\vspace{-1em}
\label{tab:sec7_xclip}
\centering
\setlength{\tabcolsep}{2pt}
\resizebox{0.78\linewidth}{!}{
\begin{tabular}{llw{c}{4em}w{c}{4em}cccccc}
\toprule
\multirow{2}{*}{Dateset} & \multirow{2}{*}{Training-strategy} & \multicolumn{2}{c}{Coarse-Grained ($\uparrow$)} &  & \multicolumn{5}{c}{Fine-Grained ($\uparrow$)} \\
\cmidrule{3-4} \cmidrule{6-10}
 &  & \multirow{1}{*}{V2T} &  \multirow{1}{*}{T2V} & & \multirow{1}{*}{noun} & \multirow{1}{*}{adj} & \multirow{1}{*}{verb} & \multirow{1}{*}{adv} & \multirow{1}{*}{prep} \\
 \midrule
\multirow{4}{*}{MSR-VTT~\cite{xu2016msr}} & Coarse-Grained Training & 66.7 & 67.0 &  & 0.430 & 0.445 & 0.314 & 0.304 & 0.225 \\
 & Word-Level Negatives & 64.5 & 67.0 &  & 0.854 & 0.818 & 0.871 & 0.675 & 0.817 \\
 & Phrase-Level Negatives & 65.6 & 67.7 &  & 0.841 & 0.795 & 0.881 & 0.512 & 0.713 \\
 & Fine-Grained Prompting &  66.2 & 67.4 &  & 0.887 & 0.897 & 0.861 & 0.723 & 0.846  \\ \midrule
 \multirow{4}{*}{VATEX~\cite{wang2019vatex}} & Coarse-Grained Training & 91.1 & 82.3 &  & 0.312 & 0.366 & 0.276 & 0.270 & 0.166   \\
 & Word-Level Negatives &85.3 & 78.2 &  & 0.875 & 0.876 & 0.911 & 0.811 & 0.720   \\
 & Phrase-Level Negatives &86.6 & 79.3 &  &0.794 & 0.757 & 0.763 & 0.672 & 0.494   \\
 & Fine-Grained Prompting & 90.3 & 81.5 &  & 0.839 & 0.774 & 0.870 & 0.738 & 0.633\\ \midrule

\multirow{4}{*}{VLN-OOPS~\cite{Voigtlaender23vln}} & Coarse-Grained Training & 70.1 & 52.3 &  & 0.409 & 0.446 & 0.236 & 0.176 & 0.142   \\
 & Word-Level Negatives &62.9 & 51.1 &  & 0.877 & 0.854 & 0.965 & 0.570 & 0.675   \\
 & Phrase-Level Negatives  &67.1 & 51.5 &  & 0.834 & 0.736 & 0.823 & 0.456 & 0.449   \\
 & Fine-Grained Prompting &67.8	&53.4 &  & 0.844 & 0.783 & 0.941 & 0.540 & 0.527  \\
 \bottomrule
\end{tabular}}
\vspace{0.3em}
\end{table}

\begin{table}[t]
\vspace{-1.0em}
\caption{\textbf{Effectiveness with Different Models}. Our approach can be combined with existing models for a large improvement in distinguishing fine-grained differences (PoSRank) and maintains coarse-grained performance (mean recall).}
\vspace{-1em}
\label{tab:sec7_uvo}
\centering
\setlength{\tabcolsep}{2pt}
\resizebox{0.78\linewidth}{!}{
\begin{tabular}{llw{c}{4em}w{c}{4em}cccccc}
\toprule
\multirow{2}{*}{Method} & \multirow{2}{*}{Training-strategy} & \multicolumn{2}{c}{Coarse-Grained ($\uparrow$)} &  & \multicolumn{5}{c}{Fine-Grained ($\uparrow$)} \\
\cmidrule{3-4} \cmidrule{6-10}
 &  & \multirow{1}{*}{V2T} &  \multirow{1}{*}{T2V} & & \multirow{1}{*}{noun} & \multirow{1}{*}{adj} & \multirow{1}{*}{verb} & \multirow{1}{*}{adv} & \multirow{1}{*}{prep} \\
 \midrule
\multirow{4}{*}{Frozen~\cite{bain2021frozen}} & Coarse-Grained Training & 38.8 & 38.6 & &  0.332 & 0.346 & 0.207 & 0.215 & 0.160 \\
 & Word-Level Negatives & 37.0 & 37.2 &  & 0.718 & 0.642 & 0.888 & 0.411 & 0.574 \\
 & Phrase-Level Negatives & 37.1 & 36.5  &  &0.770 & 0.701 & 0.887 & 0.487 & 0.627  \\
 & Fine-Grained Prompting  &39.9	&39.5&  & 0.786 & 0.733 & 0.927 & 0.490 & 0.631  \\ \midrule
 \multirow{4}{*}{UCoFiA~\cite{wang2023ucofia}} & Coarse-Grained Training & 54.4 & 39.7 &    & 0.391 & 0.423 & 0.317 & 0.227 & 0.184 \\
 & Word-Level Negatives & 43.9 & 39.0 &    & 0.894 & 0.859 & 0.973 & 0.493 & 0.696 \\
 & Phrase-Level Negatives & 50.0 & 39.1   & & 0.853 & 0.723 & 0.925 & 0.378 & 0.544  \\
 & Fine-Grained Prompting & 52.5 & 38.9 &  & 0.840 & 0.828 & 0.946 & 0.467 & 0.695 \\ \midrule
\multirow{4}{*}{TS2-Net~\cite{ts2net}} & Coarse-Grained Training &  54.1  & 39.2 & & 0.375 & 0.453 & 0.316 & 0.220 & 0.184 \\
 & Word-Level Negatives &  47.4 & 39.1 & & 0.885 & 0.852 & 0.968 & 0.461 & 0.665\\
 & Phrase-Level Negatives & 46.6& 39.2  & & 0.889 & 0.856 & 0.968 & 0.476 & 0.660 \\
 & Fine-Grained Prompting  & 54.6 & 39.7  &  &0.871 & 0.805 & 0.966 & 0.475 & 0.539   \\ \midrule
\multirow{4}{*}{X-CLIP~\cite{xclip}} & Coarse-Grained Training &   56.7 & 41.0 &  & 0.367 & 0.440 & 0.315 & 0.201 & 0.184 \\
 & Word-Level Negatives &  47.5  & 40.0 &  & 0.894 & 0.864 & 0.969 & 0.468 & 0.701 \\
 & Phrase-Level Negatives  &  52.5 & 40.2 &  & 0.839 & 0.723 & 0.913 & 0.382 & 0.527 \\
 & Fine-Grained Prompting &   52.9 & 39.8 & & 0.860 & 0.781 & 0.954 & 0.439 & 0.555  \\
 \bottomrule
\end{tabular}}
\vspace{-1em}
\end{table}

\vspace{-1.0em}
\subsection{Effectiveness with Different Models}
\vspace{-0.5em}
A benefit of our approach is that it can be easily combined with existing models to enable fine-grained video-text retrieval. We demonstrate this in \cref{tab:sec7_uvo} with results on VLN-UVO for Frozen \cite{bain2021frozen}, X-CLIP \cite{xclip}, and UCoFiA \cite{wang2023ucofia}. Our model is easy to plug into each of these methods achieving much better fine-grained performance than coarse-grained training while also maintaining or beating the original coarse-grained performance. For instance, our method with TS2-Net obtains 54.6 V2T and 0.966 verb \textit{PoSRank} vs. 54.1 and 0.316 from coarse-grained training. TS2-Net  and UCoFiA are the best-performing models on the fine-grained evaluation, however, there is no clear winner with TS2-Net performing best on nouns, verbs, and adverbs and UCoFiA performing best on adjectives and prepositions. While our approach successfully balances performance on coarse and fine-grained metrics there is still much potential for improvement in fine-grained understanding, particularly for adverbs and prepositions.  

\vspace{-1.0em}
\section{Conclusion}
\vspace{-0.5em}
In this paper, we conducted an in-depth analysis of current video retrieval evaluations finding that current benchmarks fall short in detecting a model's ability to perceive subtle single concept differences. We introduce a fine-grained approach to analyze models' sensitivity to individual word variations across different parts-of-speech. Testing this metric on current methods reveals that these models are not sensitive to such variations, especially for adverbs and prepositions. To bridge this gap, we introduce a simple, but effective baseline which can be easily combined with existing models. While experiments show that the proposed method effectively enhances the models' ability to understand fine-grained differences, there is much potential for future work.

\vspace{0.5em}
\noindent\textbf{Acknowledgements:}
\label{sec:conclusion}
We thank Yunhua Zhang for the helpful discussions, and Dennis Koelma for his invaluable technical support.

\bibliographystyle{splncs04}
\bibliography{main}
\end{document}


\title{Supplementary Materials\\ Beyond Coarse-Grained Matching in \\ Video-Text Retrieval} 

\titlerunning{Abbreviated paper title}

\author{
}


\institute{
}

\maketitle
\renewcommand\thesection{\Alph{section}}
\renewcommand\thefigure{\Alph{figure}}
\renewcommand\thetable{\Alph{table}}
\vspace{-2em}
\section{Further Analysis of the Fine-Grained Ability of Current Models}
\begin{figure}[ht]
    \centering
   
    \begin{subfigure}[b]{0.45\textwidth}
        \includegraphics[width=\textwidth]{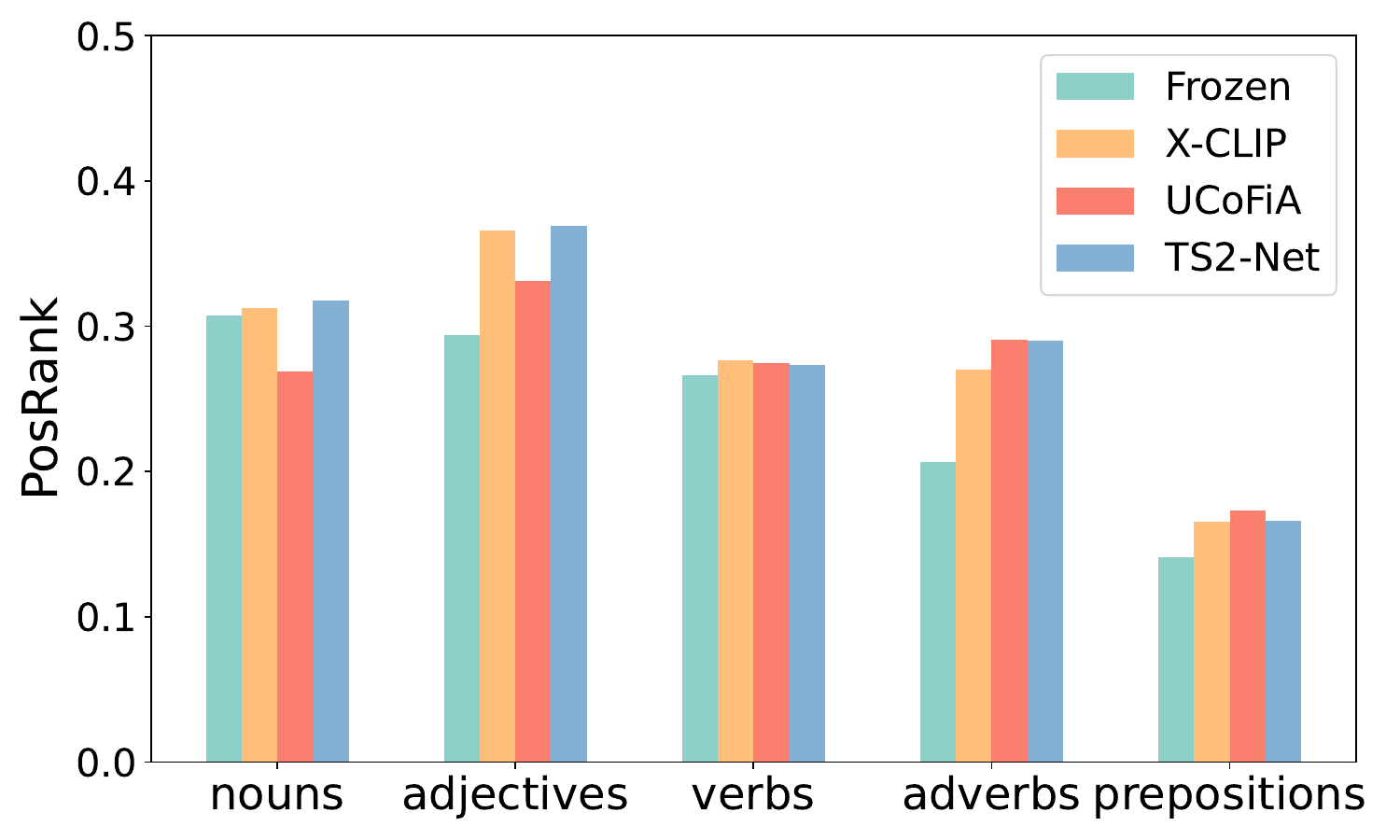}
        \caption{\textbf{VATEX}}
        \label{fig:sub1}
    \end{subfigure}
     \hspace{4mm} 
    \begin{subfigure}[b]{0.45\textwidth}
        \includegraphics[width=\textwidth]{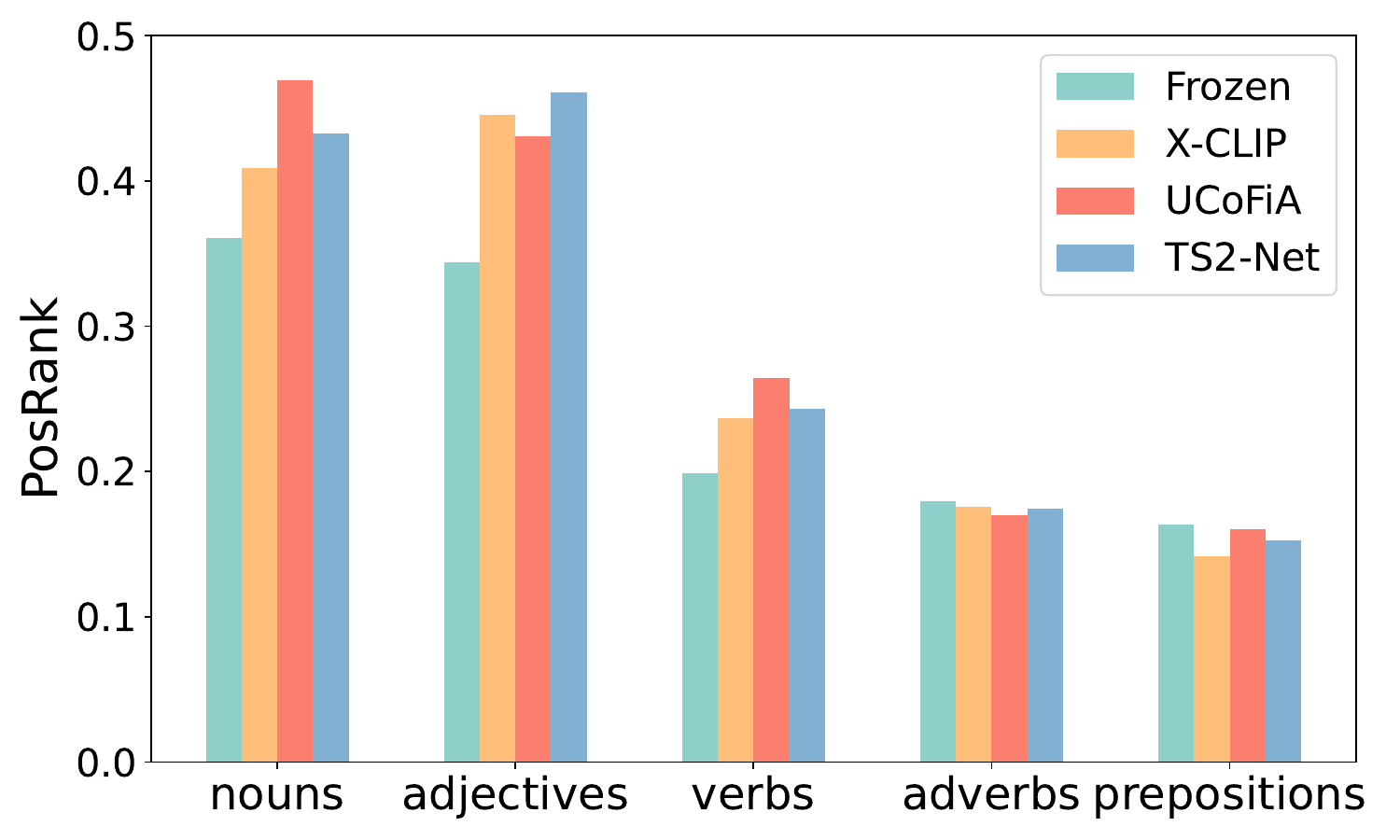}
        \caption{\textbf{VLN-OOPS}}
        \label{fig:sub3}
    \end{subfigure}

\vspace{-0.5em}
    \caption{\textbf{Fine-Grained Evaluation Per Part of Speech.} Models find fine-grained differences in adverbs and prepositions the most difficult to distinguish. }
    \vspace{-0.8em}
     \label{fig:fine_eval_bar_baseline_data}
\end{figure}

We show the \textit{PoSRank} performance for individual parts-of-speech for all methods on VATEX and VLN-OOPS in \cref{fig:fine_eval_bar_baseline_data}. We see similar trends as in Figure 6 of the main paper. Particularly, current models are better at distinguishing differences in adjectives and nouns and worse at distinguishing differences in adverbs and prepositions.

\vspace{-2em}

\section{Implementation Details of our Fine-grained Training}
X-CLIP,  TS2Net, and UCoFiA use the visual and text encoder from CLIP (ViT-B/32) \cite{2021clip_icml}, while Frozen uses a variant of ViT and DistilBERT \cite{Sanh2019distilbert} pretrained on WebVid-2M~\cite{bain2021frozen}. The training configurations are consistent with the original publications, 
All models are optimized with Adam. For X-CLIP, TS2Net and UCoFiA are trained for 5 epochs with a batch size of 64, the initial learning rate for visual and text encoder is $1e{-}7$, and the initial learning rate for other modules is $1e{-}4$. 
For Frozen, we set a learning rate of $1e{-}5$ with a maximum of 100 epochs, with early stopping if validation performance did not improve for 10 consecutive epochs, using a batch size of 32.

\section{Effect of the Number of Visual Prompts $\mathbb{T}$}

We explore the impact of the number of visual prompts in ~\cref{tab:ablation3}. With more than one visual prompt we mean-pool the embedding of all prompts for the fine-grained representation. Using any number of visual prompts $\mathbb{T}{>}0$ to better separate coarse and fine-grained objectives provides an increase in results. There is a small increase for coarse-grained evaluation with $\mathbb{T}{>}1$ however, $\mathbb{T}{=}1$ provides the best balance between coarse and fine-grained performance and training efficiency. 

\begin{table} 
\caption{\textbf{Number of visual prompts $\mathbb{T}$}.  $\mathbb{T}{=}1$ provides a reasonable balance between performance and training efficiency. }
\label{tab:ablation3}
\centering
\setlength{\tabcolsep}{5pt}
\resizebox{0.6\linewidth}{!}{ 
\begin{tabular}{llccccccc}
\toprule
\multirow{2}{*}{$\mathbb{T}$} & \multicolumn{2}{c}{Coarse-Grained ($\uparrow$)} && \multicolumn{5}{c}{Fine-Grained ($\uparrow$)} \\
\cmidrule{2-3} \cmidrule{5-9}
& \multirow{1}{*}{V2T} & \multirow{1}{*}{T2V} && \multirow{1}{*}{noun} & \multirow{1}{*}{adj} & \multirow{1}{*}{verb} & \multirow{1}{*}{adv} & \multirow{1}{*}{prep} \\
\midrule
0 & 47.5 & 40.0 && 0.894 & 0.864 & 0.969 & 0.468 & 0.701 \\
1 & 53.1 & 39.3 && 0.858 & 0.832 & 0.960 & 0.501 & 0.668 \\
2 & 54.0 & 39.8 && 0.859  &0.796 & 0.953 & 0.472 & 0.519  \\
4 & 54.0 & 40.2 &&  0.854 & 0.798 & 0.950 & 0.424 & 0.477\\
8 & 52.5 & 39.7 && 0.841 & 0.783 & 0.954 & 0.419 & 0.512 \\
\bottomrule
\end{tabular}}
\vspace{-2em}
\end{table}

\section{Variance of Results}


The table below shows the variance from 3 runs of our full model on VLU-UVO, verifying that the variance of our approach is low.

\begin{table}
\centering
\setlength{\tabcolsep}{5pt}
\resizebox{0.63\linewidth}{!}{
\begin{tabular}{cccccccc}
\toprule
\multicolumn{2}{c}{Coarse-Grained ($\uparrow$)} & &\multicolumn{5}{c}{Fine-Grained ($\uparrow$)} \\
\cmidrule{1-2} \cmidrule{4-8}
     V2T & T2V & &noun & adj & verb & adv & prep \\
     \midrule
     0.03 & 0.2 && 0.000007 & 0.000009 & 0.000001 & 0.0002 & 0.0009 \\
     \bottomrule
\end{tabular}}
\end{table}




%
%
\bibliographystyle{splncs04}
\bibliography{main}